%% file: main.tex
\documentclass{article}
\PassOptionsToPackage{numbers,compress}{natbib}
\usepackage[dblblindworkshop, final]{neurips_2026}
\workshoptitle{Grounded and Faithful Vision-Language Models for Real-World Deployment}

\usepackage[utf8]{inputenc} 
\usepackage[T1]{fontenc}    
\usepackage{hyperref}       
\usepackage{url}            
\usepackage{booktabs}       
\usepackage{amsfonts}       
\usepackage{nicefrac}       
\usepackage{microtype}      
\usepackage{xcolor}         
\usepackage{amsmath}
\usepackage{pgfplots}
\pgfplotsset{compat=1.18}
\usepgfplotslibrary{groupplots}

\usetikzlibrary{calc}

\usetikzlibrary{arrows.meta}
\usepackage{pgfplots}
\usepgfplotslibrary{groupplots}
\usetikzlibrary{calc}
\pgfplotsset{compat=1.18}

\definecolor{basecolor}{HTML}{2563EB}
\definecolor{overheadcolor}{HTML}{DC2626}

\definecolor{vblue}{RGB}{37,99,235}
\definecolor{vorange}{RGB}{217,119,6}
\definecolor{vgrid}{RGB}{220,220,220}
\definecolor{vgreen}{RGB}{0,120,0}
\definecolor{vred}{RGB}{180,0,0}
\definecolor{vgray}{RGB}{110,110,110}

\title{Towards Reliable Vision-Language Models for Autonomous Driving}

\author{%
  Manasa Mariam Mammen$^{1,2}$ \And
  Priyanka Mary Mammen$^{3}$ \And
  Zafer Kayatas$^{2}$ \And
  Stefan Wagner$^{1}$ \\[1em]
  \small
  $^{1}$Technical University of Munich, Heilbronn, Germany\\
  $^{2}$Mercedes Benz AG, Sindelfingen, Germany\\
  $^{3}$University of Massachusetts, Amherst, USA
}

\begin{document}

\maketitle

\begin{abstract}

Vision-Language models (VLMs) are increasingly being explored in autonomous driving for tasks such as scene understanding, driving reasoning, decision-making, and end-to-end driving. As their role becomes more prominent, ensuring their robustness and reliability is increasingly important. In real-world conditions, visual inputs may be degraded by sensor imperfections and environmental conditions, potentially affecting both model predictions and their associated confidence. Such degradation is especially concerning in autonomous driving, where safety-critical decisions require models to make accurate predictions and recognize when their predictions may be unreliable. In this work, we evaluate five VLMs (Qwen3.5-9B, Gemma4-E4B, LLaVA-OneVision-7B, DriveFusion/DriveFusionQA-4B, and NVIDIA Alpamayo-1.5-10B) across four driving-related QA datasets with different visual input settings, including single-frame, multi-view, multi-frame, and monocular inputs. Our results show that the effects of visual corruption vary across models, datasets, and input settings, with changes in accuracy and confidence reliability and also differing across conditions. We then apply Visual Evidence Augmentation ($\mathrm{V}{\scriptstyle \mathrm{EA}}$), a recent inference-time method to examine whether it can improve model reliability under degraded visual conditions. We find that $\mathrm{V}{\scriptstyle \mathrm{EA}}$ improves performance for some models and datasets, although the gains are not consistent across all settings.

\end{abstract}

\section{Introduction}
Autonomous driving is an emerging area in intelligent transportation systems that offers safer mobility, improved traffic efficiency, and new forms of urban transport~\cite{geiger2012we}. Traditionally, a modular pipeline architecture is followed, in which perception, prediction, planning, and control are separated into distinct subsystems~\cite{cordts2016cityscapes,dong2025end,caesar2020nuscenes}. However, these modular systems are prone to error accumulation between interfaces and struggle to achieve global optimization in highly interactive traffic environments~\cite{luo2018fast}. In response, end-to-end autonomous driving (E2E-AD) has attracted attention, in which multimodal sensory inputs are directly mapped to driving actions via unified differentiable models~\cite{bojarski2016end,yang2025survey,hu2023planning,weng2024drive}. It reduces system coupling and allows combined optimization of perception, decision-making, and control~\cite{chen2024end}. Despite the progress, the current E2E approaches still have gaps in handling long-tail and safety-critical situations, where there is a need for sparse supervision and the need high-level reasoning~\cite{wang2025alpamayo}. 

\begin{figure}[h]
    \centering
    \includegraphics[width=1\linewidth]{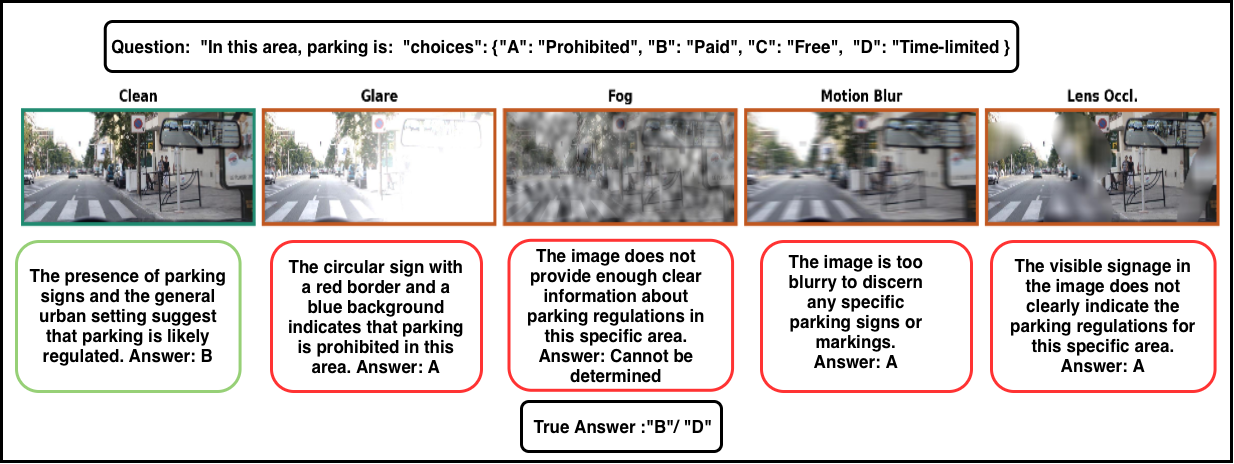}
    \caption{Gemma4-E4B responses to clean and perturbed inputs.
}
    \label{fig:gemma_example}
\end{figure}

With visual and language understanding, VLMs offer a promising direction for many parts of E2E-AD~\cite{fu2024drive,li2020traffic,mao2023gpt,mao2023language,hwang2024emma,renz2025simlingo}. They can help with tasks like perception, scene understanding, and context-aware decision-making. However, their reliability in real-world driving is still unclear. Unlike benchmark datasets, real-world driving contains many rare and difficult cases that may not be well represented in large-scale training data. For instance, in real world driving, camera inputs are rarely ideal and are affected by environmental and sensor-related challenges such as illumination changes, adverse weather etc. These perturbations/corruptions can affect not only prediction accuracy but also model confidence and calibration, which are important in safety-critical settings. As an example, Figure~\ref{fig:gemma_example} shows how Gemma4-E4B's response change when the input is perturbed compared with the clean input.

To examine the camera perturbation effects in autonomous driving, we evaluate recent vision-language models with 10B parameters or fewer (considering the importance of inference latency and computational cost for deployment). Our evaluation includes Gemma4-4B~\cite{gemmateam2026gemma4}, LLaVA-OV-8B~\cite{li2024llava}, Qwen3.5-9B~\cite{qwen3.5} , DriveFusionQA-4B~\cite{drivefusionqa2026} , and NVIDIA Alpamayo-1.5-10B~\cite{nvidia2025alpamayo}. DriveFusionQA-4B is a driving-specific VLM, while Alpamayo-1.5 is a driving-specific vision-language-action model with visual question-answering capabilities and is included because of its production-oriented design. We evaluate these models on four autonomous driving VQA benchmarks, namely DrivingVQA~\cite{corbiere2025drivingvqa} , STRIDE-QA Bench~\cite{strideqa2025} , NuScenes-QA-mini~\cite{qian2023nuscenes,huang2023modality,caesar2020nuscenes} , and Open Spatial Reasoning~\cite{driving_3d_spatial_reasoning} . These benchmarks differ in both the reasoning tasks they cover and the way visual information is presented, including single images, consecutive frames, and multi-view camera inputs. This diversity allows us to examine whether the effect of visual corruption is consistent across different driving tasks and input settings, rather than being specific to one benchmark or one data format.

To capture different sources of visual degradation, we consider four common camera perturbations, namely glare, fog, motion blur, and lens occlusion. Each one represents a broader category of real-world visual degradation. Glare represents illumination changes, fog represents adverse weather, motion blur represents sensor and motion-related effects, and lens occlusion represents partial visibility. This provides coverage of different corruption types while keeping the evaluation focused and controlled. We further investigate if $\mathrm{V}{\scriptstyle \mathrm{EA}}$~\cite{liu2026seeing}, an inference-time attention-based visual enhancement method can transfer to autonomous driving and improve model performance under corrupted inputs. This leads to the following research questions.

\paragraph{Research Questions.}
(1) How do real-world visual corruptions affect the accuracy and calibration of vision-language models across different autonomous driving tasks and visual input settings?

(2) To what extent can inference-time visual enhancement improve model robustness under real-world visual perturbations?

\section{Related Work} 

\paragraph{Corruption Robustness in VLMs }
The corruptions in  visual inputs presents a significant challenge to the reliability of VLM-based autonomous driving systems. DriveBench~\cite{xie2025vlms} was one of the first benchmarks specifically designed examining this problem, assessing VLMs under 15 different types of corruption across perception, prediction, planning, and behaviour tasks and showing that these systems have a considerable dependence on language priors when the visual input is degraded. But there has been less investigation into robustness in more diverse spatial, temporal, and complex reasoning tasks. USB~\cite{zheng2026unstructured} takes this approach further by applying it to unstructured driving situations with 20 different perturbations involving illumination, weather, occlusion, and sensor reliability, while at the same time evaluating both decision accuracy and reasoning quality. Other recent work has focused on naturally occurring adverse conditions: AWDB~\cite{awad2026benchmarking} looks at video-based scene understanding in winter driving situations, while ObsDriveBench~\cite{yan2026obsdrivebench} assesses multimodal reasoning using camera, LiDAR, and radar in real-world conditions such as fog, rain, snow, and low light. Yet despite this advancement, current robustness benchmarks generally focus on performance degradation, while uncertainty estimation and calibration  remain largely underexplored.

\paragraph{Inference-Time Methods for Improving Visual Grounding}
Several approaches have been proposed to improve how VLMs use visual information at inference time. Some methods modify the decoding process to improve visual grounding and reduce the influence of language priors, such as VCD~\cite{leng2024mitigating} and OPERA~\cite{huang2024opera}. Other methods improve visual evidence utilization through prompting or intermediate reasoning. For example, CGR~\cite{liu2025unveiling} first extracts detailed visual information and then uses this intermediate representation to generate the final answer. A growing line of work instead refines the visual input using signals from the model itself. VAR~\cite{liu2025unveiling} uses attention scores from the final layer to create a binary mask that highlights salient image regions, while AGLA~\cite{an2025mitigating} uses GradCAM-based signals to mask irrelevant regions and combines predictions from the original and masked images. $\mathrm{V}{\scriptstyle \mathrm{EA}}$~\cite{liu2026seeing} follows training-free direction by using attention-based visual enhancement to improve the use of relevant image evidence during inference. Other approaches operate on internal representations, such as VTI~\cite{liu2024reducing} and Activation Steering Decoding~\cite{su2025activation}, while fine-tuning methods such as HA-DPO~\cite{zhao2023beyond} improve visual grounding through additional training.

\section{Experimental Setup}
In this section, we describe the datasets and models, evaluation metrics, and visual corruptions used in our experiments.
\subsection{Datasets and Models}
\subsubsection*{Datasets}
We evaluate the models on four driving VQA datasets, \textbf{DrivingVQA}~\cite{corbiere2025drivingvqa}, \textbf{NuScenes-QA-mini} derived from NuScenes-QA~\cite{qian2023nuscenes}, \textbf{STRIDE-QA Bench}~\cite{strideqa2025}, and \textbf{Open Spatial Reasoning (OSR)}~\cite{driving_3d_spatial_reasoning}. DrivingVQA contains 3,931 multiple-choice questions derived from challenging real-world scenarios in French driving theory tests, with two to four answer options per question. NuScenes-QA is a multi-camera driving VQA dataset collected in Boston and Singapore, with questions covering existence, counting, object identification, object status, and comparison. We use NuScenes-QA-mini and formulate QA as open-ended text generation, rather than classification over the predefined answer set used in the original NuScenes-QA evaluation. STRIDE-QA Bench focuses on numerical and spatial reasoning in driving scenes, including distance and count estimation. OSR is a challenging monocular 3D spatial-reasoning benchmark containing 50 multiple-choice questions. Although small, we include OSR as a targeted diagnostic benchmark because it directly probes monocular 3D spatial reasoning that is less explicitly evaluated by the larger datasets. For evaluation, we use 1,000 samples from DrivingVQA and STRIDE-QA Bench, 1,117 samples from NuScenes-QA-mini, and all 50 OSR examples.

\subsubsection*{Models}
We evaluate five models including general-purpose and driving-specific vision-language architectures. The general-purpose models include \textbf{Qwen3.5-9B}, \textbf{Gemma4-E4B}, and \textbf{LLaVA-OneVision-7B}, while \textbf{DriveFusionQA-4B} is specifically fine-tuned for visual question answering and reasoning in autonomous-driving scenarios. We additionally evaluate \textbf{NVIDIA Alpamayo-1.5-10B}, a driving-specific vision-language-action model designed for autonomous driving. Although primarily developed for reasoning and trajectory prediction, Alpamayo-1.5 also explicitly supports visual question answering through its generate\_text interface, allowing it to be evaluated under the same question-answering setting. Its inclusion examine whether the observed robustness and intervention effects extend beyond conventional VLMs to models explicitly optimized for autonomous-driving reasoning.

\subsection{Evaluation Protocol and Metrics}

\subsubsection*{Dataset-specific evaluation}
The evaluation protocol follows the answer format of each benchmark. DrivingVQA and Open Spatial Reasoning (OSR) use multiple-choice questions and are evaluated using exact-match accuracy. For NuScenes-QA-mini , which contains open-ended answers, we additionally use \textbf{GPT-4} judge to account for semantically correct responses that may differ from the reference answer in wording. For STRIDE-QA Bench, we follow the benchmark's original localization criterion. A prediction is considered correct when the distance error is within $\pm 25\%$ of the ground-truth value and the heading error is within $\pm 10^\circ$~\cite{strideqa2025}.

\subsubsection*{Metrics}
We evaluate model performance using \textbf{Accuracy}, \textbf{AUROC}, \textbf{Expected Calibration Error (ECE)} and the \textbf{Brier score}. Accuracy measures task performance, while AUROC evaluates how well uncertainty distinguishes correct from incorrect predictions. For AUROC, we use \textbf{VAUQ} (Vision-Aware Uncertainty Quantification)~\cite{park2026vauq}, a training-free uncertainty score that combines predictive entropy with an Image-Information Score measuring the contribution of salient visual evidence to the prediction. Unlike entropy alone, which can be strongly influenced by language priors, VAUQ accounts for whether model confidence is supported by relevant image regions, making it well suited for visually grounded driving VQA. We further assess calibration using ECE and the Brier score. ECE measures the gap between predicted confidence and empirical accuracy, while the Brier score evaluates the overall quality of probabilistic predictions by penalizing both overconfident errors and poorly calibrated confidence estimates, with lower values indicating better calibration.

\subsection{Visual Corruptions}
For each dataset, we generate four corrupted variants of every clean image, \textbf{glare}, \textbf{fog}, \textbf{motion blur}, and \textbf{lens occlusion}, covering common illumination, weather, motion, and occlusion degradations. Following the Unstructured Scene Benchmark (USB)~\cite{zheng2026unstructured}, corruptions are applied independently using a fixed random seed, with severity parameters sampled per image from predefined ranges.

\textbf{Glare.} We overlay a Gaussian glare pattern centered at a random image location, with radius $r \sim \mathcal{U}(150,300)$ and intensity $\alpha \sim \mathcal{U}(1.0,1.5)$,
\begin{equation}
I'=\mathrm{clip}!\left(I+\alpha \cdot 255
\exp!\left(-\frac{(x-c_x)^2+(y-c_y)^2}{2r^2}\right),0,255\right).
\end{equation}

\textbf{Fog.} A four-octave smoothed noise mask $M\in[0,1]^{H\times W}$ models spatially varying haze, with opacity $\alpha\sim\mathcal{U}(0.4,0.8)$,
\begin{equation}
I'=(1-\alpha)I+\alpha\cdot255\cdot M.
\end{equation}

\textbf{Motion blur.} We convolve the image with a horizontal linear-motion kernel of uniformly sampled integer size $k\in{15,\ldots,34}$,
\begin{equation}
I'=I*K,
\end{equation}
where the central row of $K$ has entries $1/k$ and all other entries are zero.

\textbf{Lens occlusion.} We place $n\in{2,3,4}$ randomly positioned and oriented elliptical regions and replace their contents with a strongly blurred version of the image,
\begin{equation}
I'=I\odot(1-M)+\mathrm{Blur}_{51,50}(I)\odot M,
\end{equation}
where $M$ is the union of the sampled ellipses.

\section{Results}
In this section, we first examine how visual corruptions affect model performance and confidence, and then evaluate whether $\mathrm{V}{\scriptstyle \mathrm{EA}}$ improves model performance while maintaining reliable confidence under degraded visual inputs.

\subsection{Robustness Analysis}
We first analyze model robustness under the four representative corruption conditions using accuracy, AUROC$_{\mathrm{VAUQ}}$, ECE, and Brier score, with each corrupted compared against the corresponding clean input. Table~\ref{tab:capability-by-dataset-drivingvqa} reports the results on DrivingVQA, and Table~\ref{tab:capability-by-dataset-nuscenes-qa} reports the results on NuScenes-QA-mini. Results on STRIDE-QA Bench and Open Spatial Reasoning (OSR) are given in the appendix in Tables~\ref{tab:capability-by-dataset-stride-qa} and~\ref{tab:capability-by-dataset-osr}.

On DrivingVQA, where questions combine driving scenes with driving knowledge, accuracy changes only slightly for most of the models. However, the confidence metrics show a different picture here. DriveFusionQA-4B stays close to chance AUROC$_{\mathrm{VAUQ}}$ even though it has relatively low ECE and Brier scores, suggesting that its confidence may be reasonable on average but is not very useful for separating correct from incorrect predictions. Qwen3.5-9B, Gemma4-E4B, and Alpamayo-1.5-10B show more above-chance discrimination in several settings, although this does not always match their accuracy or calibration. Gemma4-E4B is a clear example where its accuracy changes only moderately under some corruptions, while AUROC$_{\mathrm{VAUQ}}$ drops much more. 

\providecolor{vgreen}{RGB}{0,120,0}
\providecolor{vred}{RGB}{180,0,0}


\begin{table}[h]
\centering
\small

\caption{DrivingVQA results under clean and corrupted conditions. We report Accuracy, AUROC$_{\mathrm{VAUQ}}$, ECE, and Brier score. \textcolor{vgreen}{Green} and \textcolor{vred}{red} show improvement and degradation relative to clean performance.}

\label{tab:capability-by-dataset-drivingvqa}

\resizebox{\linewidth}{!}{%
\begin{tabular}{l|rrrr|rrrr|rrrr}
\toprule
& \multicolumn{4}{c}{Qwen3.5-9B}
& \multicolumn{4}{c}{DriveFusionQA-4B}
& \multicolumn{4}{c}{Gemma4-E4B} \\
\cmidrule(lr){2-5}
\cmidrule(lr){6-9}
\cmidrule(lr){10-13}

Condition
& Acc  $\uparrow$ & AUROC  $\uparrow$ & ECE $\downarrow$& Brier $\downarrow$
& Acc  $\uparrow$ & AUROC  $\uparrow$ & ECE $\downarrow$& Brier $\downarrow$
& Acc  $\uparrow$ & AUROC  $\uparrow$ & ECE $\downarrow$& Brier $\downarrow$ \\
\midrule

Clean
& 55.8 & 53.6 & 21.5 & 29.3
& 50.4 & 49.0 & 13.5 & 27.2
& 55.2 & 55.1 & 19.5 & 28.4 \\

Glare
& \textcolor{vgreen}{56.4} & \textcolor{vgreen}{54.6} & \textcolor{vgreen}{19.8} & \textcolor{vgreen}{28.2}
& \textcolor{vgreen}{50.9} & \textcolor{vred}{48.7} & \textcolor{vgreen}{12.3} & \textcolor{vgreen}{26.9}
& \textcolor{vred}{54.7} & \textcolor{vred}{52.3} & \textcolor{vred}{20.2} & \textcolor{vred}{28.8} \\

Fog
& \textcolor{vred}{55.3} & \textcolor{vred}{52.2} & \textcolor{vred}{21.9} & \textcolor{vred}{29.5}
& \textcolor{vgreen}{52.2} & \textcolor{vgreen}{49.4} & \textcolor{vgreen}{11.2} & \textcolor{vgreen}{26.6}
& \textcolor{vred}{53.6} & \textcolor{vred}{52.8} & \textcolor{vred}{21.3} & \textcolor{vred}{29.5} \\

Motion Blur
& \textcolor{vred}{55.3} & \textcolor{vred}{53.5} & \textcolor{vgreen}{21.3} & \textcolor{vgreen}{29.3}
& \textcolor{vgreen}{50.7} & \textcolor{vgreen}{51.4} & \textcolor{vgreen}{10.9} & \textcolor{vgreen}{26.5}
& \textcolor{vred}{53.3} & \textcolor{vred}{49.6} & \textcolor{vred}{20.8} & \textcolor{vred}{29.8} \\

Lens Occl.
& \textcolor{vred}{53.8} & \textcolor{vred}{52.3} & \textcolor{vred}{22.6} & \textcolor{vred}{30.1}
& \textcolor{vgreen}{50.9} & \textcolor{vgreen}{51.0} & \textcolor{vgreen}{11.2} & \textcolor{vgreen}{26.7}
& \textcolor{vred}{53.5} & \textcolor{vred}{52.3} & \textcolor{vred}{21.9} & \textcolor{vred}{29.7} \\

\bottomrule
\end{tabular}%
}

\vspace{0.8em}

\begin{tabular}{l|rrrr|rrrr}
\toprule
& \multicolumn{4}{c}{LLaVA-OV-7B}
& \multicolumn{4}{c}{Alpamayo} \\
\cmidrule(lr){2-5}
\cmidrule(lr){6-9}

Condition
& Acc  $\uparrow$ & AUROC  $\uparrow$ & ECE $\downarrow$& Brier $\downarrow$
& Acc  $\uparrow$ & AUROC  $\uparrow$ & ECE $\downarrow$& Brier $\downarrow$ \\
\midrule

Clean
& 53.8 & 51.3 & 21.9 & 30.2
& 45.5 & 56.6 & 24.0 & 30.8 \\

Glare
& \textcolor{vred}{52.8} & \textcolor{vgreen}{52.4} & \textcolor{vgreen}{21.4} & \textcolor{vgreen}{29.9}
& \textcolor{vred}{45.0} & \textcolor{vred}{56.1} & \textcolor{vred}{24.4} & \textcolor{vred}{31.3} \\

Fog
& \textcolor{vred}{51.6} & \textcolor{vgreen}{52.4} & \textcolor{vred}{22.4} & \textcolor{vred}{30.6}
& 45.5 & \textcolor{vgreen}{57.3} & \textcolor{vred}{24.4} & \textcolor{vred}{31.1} \\

Motion Blur
& 53.8 & \textcolor{vred}{50.8} & \textcolor{vgreen}{19.7} & \textcolor{vgreen}{29.4}
& \textcolor{vred}{44.4} & \textcolor{vred}{56.1} & \textcolor{vred}{25.1} & \textcolor{vred}{31.8} \\

Lens Occl.
& \textcolor{vred}{53.2} & \textcolor{vgreen}{53.0} & \textcolor{vgreen}{21.1} & \textcolor{vgreen}{29.8}
& \textcolor{vred}{44.3} & \textcolor{vred}{55.7} & \textcolor{vred}{25.2} & \textcolor{vred}{31.6} \\

\bottomrule
\end{tabular}

\end{table}
On NuScenes-QA-mini, which relies more on object-level, counting, directional, and relational reasoning, the effect of corruption is stronger. Qwen3.5-9B, Gemma4-E4B, and LLaVA-OV-7B retain above-chance AUROC$_{\mathrm{VAUQ}}$ under several corruptions, suggesting that some visual information is still being used, although the drop from the clean setting shows that this signal becomes less reliable. DriveFusionQA-4B shows a different pattern. Its accuracy is slightly higher under the corrupted conditions and its ECE decreases, while AUROC$_{\mathrm{VAUQ}}$ consistently drops. This suggests that the small gain in task accuracy does not reflect better visually grounded reliability.

\begin{table}[h]
\centering
\small

\caption{NuScenes-QA-mini results under clean and corrupted conditions. We report Accuracy, AUROC$_{\mathrm{VAUQ}}$, ECE, and Brier score. \textcolor{vgreen}{Green} and \textcolor{vred}{red} show improvement and degradation relative to clean performance.
}
\label{tab:capability-by-dataset-nuscenes-qa}

\resizebox{\linewidth}{!}{%
\begin{tabular}{l|rrrr|rrrr|rrrr}
\toprule
& \multicolumn{4}{c}{Qwen3.5-9B}
& \multicolumn{4}{c}{DriveFusionQA-4B}
& \multicolumn{4}{c}{Gemma4-E4B} \\
\cmidrule(lr){2-5}
\cmidrule(lr){6-9}
\cmidrule(lr){10-13}

Condition
& Acc  $\uparrow$ & AUROC  $\uparrow$ & ECE $\downarrow$& Brier $\downarrow$
& Acc  $\uparrow$ & AUROC  $\uparrow$ & ECE $\downarrow$& Brier $\downarrow$
& Acc  $\uparrow$ & AUROC  $\uparrow$ & ECE $\downarrow$& Brier $\downarrow$\\
\midrule

Clean
& 42.1 & 67.0 & 49.5 & 47.5
& 35.3 & 61.6 & 23.8 & 27.2
& 38.0 & 63.5 & 54.7 & 51.7 \\

Glare
& \textcolor{vred}{29.5} & \textcolor{vred}{60.0} & \textcolor{vred}{63.4} & \textcolor{vred}{60.0}
& \textcolor{vgreen}{36.4} & \textcolor{vred}{54.9} & \textcolor{vgreen}{22.0} & \textcolor{vred}{28.5}
& \textcolor{vred}{29.8} & \textcolor{vred}{55.1} & \textcolor{vred}{63.3} & \textcolor{vred}{60.1} \\

Fog
& \textcolor{vred}{40.6} & \textcolor{vred}{61.9} & \textcolor{vred}{52.2} & \textcolor{vred}{50.5}
& \textcolor{vgreen}{37.7} & \textcolor{vred}{59.7} & \textcolor{vgreen}{21.6} & \textcolor{vred}{28.1}
& \textcolor{vred}{35.2} & \textcolor{vred}{59.3} & \textcolor{vred}{58.0} & \textcolor{vred}{55.0} \\

Motion Blur
& \textcolor{vred}{41.5} & \textcolor{vred}{61.8} & \textcolor{vred}{51.9} & \textcolor{vred}{50.4}
& \textcolor{vgreen}{36.3} & \textcolor{vred}{58.3} & \textcolor{vgreen}{23.1} & \textcolor{vred}{28.3}
& \textcolor{vred}{34.6} & \textcolor{vred}{57.8} & \textcolor{vred}{59.5} & \textcolor{vred}{56.8} \\

Lens Occl.
& \textcolor{vred}{40.9} & \textcolor{vred}{65.4} & \textcolor{vred}{51.2} & \textcolor{vred}{49.4}
& \textcolor{vgreen}{37.2} & \textcolor{vred}{59.3} & \textcolor{vgreen}{23.3} & \textcolor{vred}{28.1}
& \textcolor{vred}{36.3} & \textcolor{vred}{62.4} & \textcolor{vred}{57.1} & \textcolor{vred}{54.0} \\

\bottomrule
\end{tabular}%
}

\vspace{0.8em}

\begin{tabular}{l|rrrr|rrrr}
\toprule
& \multicolumn{4}{c}{LLaVA-OV-7B}
& \multicolumn{4}{c}{Alpamayo} \\
\cmidrule(lr){2-5}
\cmidrule(lr){6-9}

Condition
& Acc  $\uparrow$ & AUROC  $\uparrow$ & ECE $\downarrow$& Brier $\downarrow$
& Acc  $\uparrow$ & AUROC  $\uparrow$ & ECE $\downarrow$& Brier $\downarrow$ \\
\midrule

Clean
& 44.5 & 64.7 & 29.5 & 30.6
& 50.0 & 51.7 & 19.1 & 28.6 \\

Glare
& \textcolor{vred}{37.2} & \textcolor{vred}{52.1} & \textcolor{vred}{40.4} & \textcolor{vred}{40.6}
& \textcolor{vred}{45.1} & \textcolor{vgreen}{55.2} & \textcolor{vred}{21.9} & \textcolor{vred}{29.4} \\

Fog
& \textcolor{vred}{42.3} & \textcolor{vred}{64.2} & \textcolor{vred}{32.3} & \textcolor{vred}{32.2}
& \textcolor{vred}{47.4} & \textcolor{vgreen}{52.3} & \textcolor{vred}{19.8} & \textcolor{vred}{28.8} \\

Motion Blur
& \textcolor{vred}{39.8} & \textcolor{vred}{60.1} & \textcolor{vred}{33.4} & \textcolor{vred}{33.6}
& \textcolor{vred}{44.2} & \textcolor{vgreen}{54.3} & \textcolor{vred}{20.6} & \textcolor{vred}{28.9} \\

Lens Occl.
& \textcolor{vgreen}{44.7} & \textcolor{vred}{60.2} & \textcolor{vred}{29.8} & \textcolor{vred}{31.2}
& \textcolor{vred}{43.6} & \textcolor{vgreen}{60.5} & \textcolor{vred}{20.4} & \textcolor{vgreen}{28.1} \\

\bottomrule
\end{tabular}

\end{table}

Compared with \cite{xie2025vlms} and \cite{zheng2026unstructured}, which show that VLMs can still give plausible answers when the visual input is degraded, our results suggest that models can still make some use of the corrupted visual input. However, their confidence does not always reflect when this visual evidence leads to a correct prediction. AUROC$_{\mathrm{VAUQ}}$ can decrease even when accuracy changes only slightly, showing that stable task performance does not necessarily mean reliable visually grounded reasoning. We also find that no model performs consistently well across accuracy, calibration, and AUROC$_{\mathrm{VAUQ}}$. Some models are better calibrated but have weak discrimination between correct and incorrect predictions, while others show stronger discrimination but poor calibration. These results point to a reliability gap in current VLMs for driving tasks, where stable task performance does not always mean that the model can reliably judge whether its answer is supported by the visual input.

\subsection{Visual Evidence Enhancement}

Table~\ref{tab:vea-all-datasets} summarizes the effect of $\mathrm{V}{\scriptstyle \mathrm{EA}}$ on accuracy across all four datasets. We use these results to examine whether $\mathrm{V}{\scriptstyle \mathrm{EA}}$ helps on clean inputs and whether any gains remain under visual corruptions. Overall, its effect depends on the model and task. In some models, improvements on clean inputs are maintained under perturbations, as with Qwen3.5-9B on DrivingVQA and STRIDE-QA Bench and Alpamayo-1.5-10B on DrivingVQA and NuScenes-QA-mini. When $\mathrm{V}{\scriptstyle \mathrm{EA}}$ is not helpful on clean inputs, the same limitation often remains under corruption, as seen for LLaVA-OV-7B on NuScenes-QA-mini and Alpamayo-1.5-10B on STRIDE-QA Bench. This suggests that $\mathrm{V}{\scriptstyle \mathrm{EA}}$ can remain effective under degraded visual inputs, but not consistently across all model and dataset combinations. 

STRIDE-QA Bench also shows an important limitation. $\mathrm{V}{\scriptstyle \mathrm{EA}}$ re-weights spatial evidence but does not add temporal motion information, so it cannot directly address reasoning that depends strongly on temporal cues. This explains why Alpamayo-1.5-10B, despite benefiting strongly on DrivingVQA and NuScenes-QA-mini and mostly improving on OSR, consistently loses accuracy on STRIDE-QA Bench.

\begin{table}[h]
\centering
\small

\caption{Change in accuracy after applying $\mathrm{V}{\scriptstyle \mathrm{EA}}$ across all four datasets.
\textcolor{vgreen}{Green} and \textcolor{vred}{red} indicate improvement
and degradation, respectively.}
\label{tab:vea-all-datasets}

\resizebox{\linewidth}{!}{%
\begin{tabular}{l*{12}{r}}
\toprule
& \multicolumn{4}{c}{Qwen3.5-9B}
& \multicolumn{4}{c}{DriveFusionQA-4B}
& \multicolumn{4}{c}{Gemma4-E4B} \\
\cmidrule(lr){2-5}
\cmidrule(lr){6-9}
\cmidrule(lr){10-13}

Condition
& DVQA & NQA & OSR & STRIDE
& DVQA & NQA & OSR & STRIDE
& DVQA & NQA & OSR & STRIDE \\
\midrule

Clean
& \textcolor{vgreen}{+3.20}
& \textcolor{vred}{-0.18}
& \textcolor{vgreen}{+8.00}
& \textcolor{vgreen}{+6.00}
& \textcolor{vgreen}{+1.30}
& \textcolor{vgreen}{+1.25}
& +0.00
& \textcolor{vred}{-0.90}
& \textcolor{vgreen}{+0.80}
& \textcolor{vred}{-0.81}
& \textcolor{vred}{-12.00}
& \textcolor{vred}{-1.10} \\

Glare
& \textcolor{vgreen}{+0.70}
& \textcolor{vgreen}{+5.37}
& \textcolor{vred}{-10.00}
& \textcolor{vgreen}{+8.60}
& \textcolor{vgreen}{+0.10}
& \textcolor{vgreen}{+0.72}
& \textcolor{vgreen}{+2.00}
& \textcolor{vred}{-1.60}
& \textcolor{vred}{-0.60}
& \textcolor{vred}{-0.54}
& \textcolor{vred}{-14.00}
& \textcolor{vgreen}{+0.80} \\

Fog
& \textcolor{vgreen}{+2.20}
& \textcolor{vgreen}{+0.27}
& \textcolor{vred}{-2.00}
& \textcolor{vgreen}{+6.50}
& \textcolor{vred}{-0.40}
& \textcolor{vred}{-1.43}
& \textcolor{vred}{-2.00}
& \textcolor{vgreen}{+1.30}
& \textcolor{vgreen}{+1.70}
& \textcolor{vred}{-0.09}
& \textcolor{vred}{-8.00}
& \textcolor{vgreen}{+1.60} \\

Motion Blur
& \textcolor{vgreen}{+1.50}
& \textcolor{vred}{-0.27}
& \textcolor{vgreen}{+4.00}
& \textcolor{vgreen}{+5.50}
& \textcolor{vgreen}{+0.20}
& \textcolor{vgreen}{+0.90}
& +0.00
& \textcolor{vgreen}{+1.40}
& \textcolor{vgreen}{+3.00}
& \textcolor{vred}{-0.54}
& \textcolor{vgreen}{+2.00}
& \textcolor{vred}{-2.40} \\

Lens Occl.
& \textcolor{vgreen}{+5.00}
& \textcolor{vgreen}{+0.54}
& \textcolor{vgreen}{+8.00}
& \textcolor{vgreen}{+6.70}
& \textcolor{vred}{-0.10}
& \textcolor{vgreen}{+0.54}
& +0.00
& \textcolor{vgreen}{+0.30}
& \textcolor{vgreen}{+2.80}
& \textcolor{vgreen}{+0.09}
& \textcolor{vgreen}{+8.00}
& \textcolor{vgreen}{+0.50} \\

\bottomrule
\end{tabular}%
}

\vspace{0.8em}

\begin{tabular}{l*{8}{r}}
\toprule
& \multicolumn{4}{c}{LLaVA-OV-7B}
& \multicolumn{4}{c}{Alpamayo-1.5-10B} \\
\cmidrule(lr){2-5}
\cmidrule(lr){6-9}

Condition
& DVQA & NQA & OSR & STRIDE
& DVQA & NQA & OSR & STRIDE \\
\midrule

Clean
& \textcolor{vgreen}{+0.50}
& \textcolor{vred}{-4.03}
& \textcolor{vred}{-4.00}
& \textcolor{vgreen}{+3.30}
& \textcolor{vgreen}{+7.40}
& \textcolor{vgreen}{+7.97}
& \textcolor{vgreen}{+4.44}
& \textcolor{vred}{-4.10} \\

Glare
& +0.00
& \textcolor{vred}{-2.42}
& \textcolor{vred}{-6.00}
& \textcolor{vgreen}{+1.70}
& \textcolor{vgreen}{+6.70}
& \textcolor{vgreen}{+3.22}
& +0.00
& \textcolor{vred}{-3.00} \\

Fog
& \textcolor{vgreen}{+1.40}
& \textcolor{vred}{-2.42}
& \textcolor{vred}{-6.00}
& \textcolor{vgreen}{+1.10}
& \textcolor{vgreen}{+7.80}
& \textcolor{vgreen}{+4.39}
& \textcolor{vgreen}{+9.09}
& \textcolor{vred}{-1.70} \\

Motion Blur
& \textcolor{vgreen}{+0.10}
& \textcolor{vred}{-1.16}
& +0.00
& \textcolor{vgreen}{+2.80}
& \textcolor{vgreen}{+7.50}
& \textcolor{vgreen}{+7.34}
& \textcolor{vgreen}{+4.55}
& \textcolor{vred}{-2.00} \\

Lens Occl.
& \textcolor{vred}{-0.40}
& \textcolor{vred}{-3.76}
& \textcolor{vgreen}{+2.00}
& \textcolor{vgreen}{+3.20}
& \textcolor{vgreen}{+7.60}
& \textcolor{vgreen}{+8.50}
& \textcolor{vgreen}{+2.17}
& \textcolor{vred}{-1.30} \\

\bottomrule
\end{tabular}

\end{table}

We then examine whether these performance gains are accompanied by more reliable confidence estimates. Figure~\ref{fig:vea-lines} compares accuracy, AUROC$_{\mathrm{VAUQ}}$, ECE, and Brier before and after $\mathrm{V}{\scriptstyle \mathrm{EA}}$ on DrivingVQA. The results show that higher accuracy does not always come with more reliable confidence. Qwen3.5-9B and Gemma4-E4B generally improve in accuracy and calibration, although their AUROC$_{\mathrm{VAUQ}}$ changes are mixed. DriveFusionQA-4B gains little in accuracy, but its AUROC$_{\mathrm{VAUQ}}$ and calibration metrics generally improve, suggesting that $\mathrm{V}{\scriptstyle \mathrm{EA}}$ provides a larger benefit to reliability than to task performance. In contrast, Alpamayo-1.5-10B shows large and consistent accuracy gains while AUROC$_{\mathrm{VAUQ}}$, ECE, and Brier worsen, showing that better task performance does not necessarily mean that confidence becomes more reliable. LLaVA-OV-7B also shows a trade-off between correctness discrimination and calibration.

Overall, $\mathrm{V}{\scriptstyle \mathrm{EA}}$ can improve performance and remains useful even under corruption for some models. But these gains are not always accompanied by more reliable confidence. The corresponding results for NuScenes-QA-mini Figure~\ref{fig:vea_nuscenes}, STRIDE-QA Bench Figure~\ref{fig:vea_stride}, and OSR Figure~\ref{fig:vea_osr} are provided in the appendix.

\input{line_vea}

\section{Conclusion}
In this work, we evaluated general purpose and driving specialized VLMs under clean and corrupted visual conditions using accuracy and confidence. We found that model performance and confidence reliability do not always change in the same way, and no model performs consistently well across accuracy, calibration, and AUROC. We also evaluated $\mathrm{V}{\scriptstyle \mathrm{EA}}$ as an inference time method and found that it improves performance for some model and dataset combinations, with many of these gains remaining under corruption. However, its effect is task and model dependent, and improved accuracy does not always come with more reliable confidence.

\bibliographystyle{unsrtnat}
\bibliography{references}
\appendix
\section{Supplementary Results}
For STRIDE-QA Bench, Table~\ref{tab:capability-by-dataset-stride-qa} shows the lowest overall accuracy among the four datasets, which reflects the difficulty of its spatiotemporal reasoning setting. Under corruption, the changes are mixed and should not be interpreted as clear gains or losses from a single metric. Qwen3.5-9B and DriveFusionQA-4B often show lower AUROC$_{\mathrm{VAUQ}}$, while Gemma4-E4B and Alpamayo-1.5-10B sometimes show better ECE or Brier despite limited or lower accuracy. LLaVA-OV-7B remains below chance in AUROC$_{\mathrm{VAUQ}}$ across all conditions, suggesting that its confidence is not reliably aligned with visually grounded correctness in this setting.


\begin{table}[h]
\centering
\small

\caption{STRIDE-QA Bench results under clean and corrupted conditions. We report Accuracy, AUROC$_{\mathrm{VAUQ}}$, ECE, and Brier score. \textcolor{vgreen}{Green} and \textcolor{vred}{red} show improvement and degradation relative to clean performance.}
\label{tab:capability-by-dataset-stride-qa}

\resizebox{\linewidth}{!}{%
\begin{tabular}{l|rrrr|rrrr|rrrr}
\toprule
& \multicolumn{4}{c}{Qwen3.5-9B}
& \multicolumn{4}{c}{DriveFusionQA-4B}
& \multicolumn{4}{c}{Gemma4-E4B} \\
\cmidrule(lr){2-5}
\cmidrule(lr){6-9}
\cmidrule(lr){10-13}

Condition
& Acc  $\uparrow$ & AUROC  $\uparrow$ & ECE $\downarrow$& Brier $\downarrow$
& Acc  $\uparrow$ & AUROC  $\uparrow$ & ECE $\downarrow$& Brier $\downarrow$
& Acc  $\uparrow$ & AUROC  $\uparrow$ & ECE $\downarrow$& Brier $\downarrow$ \\
\midrule

Clean
& 18.1 & 51.1 & 53.1 & 43.3
& 8.9 & 54.4 & 31.2 & 18.1
& 13.1 & 53.1 & 61.7 & 49.8 \\

Glare
& \textcolor{vred}{15.9} & \textcolor{vred}{49.3} & \textcolor{vred}{55.4} & \textcolor{vred}{44.3}
& \textcolor{vgreen}{9.4} & \textcolor{vred}{51.5} & \textcolor{vgreen}{30.1} & \textcolor{vgreen}{17.8}
& \textcolor{vred}{12.1} & \textcolor{vgreen}{55.1} & \textcolor{vred}{62.3} & \textcolor{vgreen}{49.7} \\

Fog
& \textcolor{vred}{17.7} & \textcolor{vred}{43.9} & \textcolor{vred}{54.1} & \textcolor{vred}{44.3}
& \textcolor{vred}{7.6} & \textcolor{vred}{50.8} & \textcolor{vred}{31.7} & \textcolor{vgreen}{17.2}
& \textcolor{vgreen}{14.6} & \textcolor{vred}{49.5} & \textcolor{vgreen}{58.9} & \textcolor{vgreen}{47.6} \\

Motion Blur
& \textcolor{vgreen}{19.9} & \textcolor{vred}{46.2} & \textcolor{vgreen}{51.8} & \textcolor{vgreen}{43.1}
& \textcolor{vred}{7.7} & \textcolor{vred}{46.8} & \textcolor{vred}{31.4} & \textcolor{vgreen}{17.5}
& \textcolor{vgreen}{16.7} & \textcolor{vgreen}{55.0} & \textcolor{vgreen}{58.3} & \textcolor{vgreen}{48.2} \\

Lens Occl.
& \textcolor{vred}{16.8} & \textcolor{vred}{49.3} & \textcolor{vred}{54.4} & \textcolor{vred}{43.7}
& \textcolor{vred}{8.5} & \textcolor{vred}{50.1} & \textcolor{vgreen}{30.9} & \textcolor{vgreen}{17.6}
& \textcolor{vred}{12.8} & \textcolor{vgreen}{54.8} & \textcolor{vgreen}{61.5} & \textcolor{vgreen}{49.2} \\
\hline

\end{tabular}%
}

\vspace{0.8em}

\begin{tabular}{l|rrrr|rrrr}
\toprule
& \multicolumn{4}{c}{LLaVA-OV-7B}
& \multicolumn{4}{c}{Alpamayo} \\
\cmidrule(lr){2-5}
\cmidrule(lr){6-9}

Condition
& Acc & AUROC & ECE & Brier
& Acc & AUROC & ECE & Brier \\
\midrule

Clean
& 11.7 & 33.6 & 39.5 & 29.4
& 9.5 & 53.0 & 56.6 & 43.0 \\

Glare
& \textcolor{vgreen}{12.8} & \textcolor{vgreen}{37.1} & \textcolor{vgreen}{38.3} & \textcolor{vgreen}{29.1}
& \textcolor{vred}{8.6} & \textcolor{vred}{52.3} & \textcolor{vgreen}{52.0} & \textcolor{vgreen}{37.2} \\

Fog
& \textcolor{vred}{11.0} & \textcolor{vgreen}{34.7} & \textcolor{vred}{40.2} & \textcolor{vgreen}{29.2}
& \textcolor{vred}{7.2} & \textcolor{vgreen}{54.4} & \textcolor{vgreen}{50.0} & \textcolor{vgreen}{33.7} \\

Motion Blur
& \textcolor{vred}{11.4} & \textcolor{vgreen}{41.0} & \textcolor{vred}{39.8} & \textcolor{vgreen}{28.5}
& \textcolor{vred}{7.5} & \textcolor{vred}{50.9} & \textcolor{vgreen}{50.8} & \textcolor{vgreen}{34.8} \\

Lens Occl.
& \textcolor{vred}{11.0} & \textcolor{vgreen}{37.5} & \textcolor{vred}{40.6} & \textcolor{vgreen}{29.0}
& \textcolor{vred}{6.9} & \textcolor{vgreen}{57.3} & \textcolor{vgreen}{50.1} & \textcolor{vgreen}{33.5} \\

\bottomrule
\end{tabular}

\end{table}


For OSR, Table~\ref{tab:capability-by-dataset-osr} also shows strong disagreement between accuracy, AUROC$_{\mathrm{VAUQ}}$, and calibration under corruption. Some conditions produce higher accuracy or lower calibration error, while AUROC$_{\mathrm{VAUQ}}$ moves in the opposite direction, showing that these changes do not necessarily reflect more reliable reasoning. Since OSR contains only 50 samples, the condition level differences are particularly noisy and are best treated as supporting evidence rather than firm conclusions.

\begin{table}[h]
\centering
\small

\caption{OSR results under clean and corrupted conditions. We report Accuracy, AUROC$_{\mathrm{VAUQ}}$, ECE, and Brier score. \textcolor{vgreen}{Green} and \textcolor{vred}{red} show improvement and degradation relative to clean performance.}
\label{tab:capability-by-dataset-osr}

\resizebox{\linewidth}{!}{%
\begin{tabular}{l|rrrr|rrrr|rrrr}
\toprule
& \multicolumn{4}{c}{Qwen3.5-9B}
& \multicolumn{4}{c}{DriveFusionQA-4B}
& \multicolumn{4}{c}{Gemma4-E4B} \\
\cmidrule(lr){2-5}
\cmidrule(lr){6-9}
\cmidrule(lr){10-13}

Condition
& Acc  $\uparrow$ & AUROC  $\uparrow$ & ECE $\downarrow$& Brier $\downarrow$
& Acc  $\uparrow$ & AUROC  $\uparrow$ & ECE $\downarrow$& Brier $\downarrow$
& Acc  $\uparrow$ & AUROC $\uparrow$  & ECE$\downarrow$ & Brier $\downarrow$\\
\midrule

Clean
& 42.0 & 53.4 & 27.4 & 32.1
& 20.0 & 44.5 & 54.6 & 46.9
& 42.0 & 56.7 & 34.4 & 36.7 \\

Glare
& \textcolor{vgreen}{50.0} & \textcolor{vgreen}{63.0} & \textcolor{vgreen}{19.3} & \textcolor{vgreen}{27.8}
& \textcolor{vgreen}{22.0} & \textcolor{vgreen}{45.0} & \textcolor{vgreen}{52.7} & \textcolor{vgreen}{46.2}
& \textcolor{vgreen}{46.0} & \textcolor{vred}{42.5} & \textcolor{vgreen}{31.2} & \textcolor{vgreen}{35.9} \\

Fog
& \textcolor{vgreen}{48.0} & \textcolor{vgreen}{57.4} & \textcolor{vgreen}{21.8} & \textcolor{vgreen}{29.1}
& \textcolor{vgreen}{28.0} & \textcolor{vred}{42.9} & \textcolor{vgreen}{45.8} & \textcolor{vgreen}{43.1}
& 42.0 & \textcolor{vred}{41.2} & \textcolor{vred}{34.7} & \textcolor{vred}{38.0} \\

Motion Blur
& \textcolor{vred}{36.0} & \textcolor{vgreen}{63.2} & \textcolor{vred}{33.5} & \textcolor{vred}{33.1}
& \textcolor{vgreen}{22.0} & \textcolor{vred}{36.1} & \textcolor{vgreen}{50.9} & \textcolor{vgreen}{44.2}
& \textcolor{vred}{36.0} & \textcolor{vred}{38.2} & \textcolor{vred}{40.9} & \textcolor{vred}{40.9} \\

Lens Occl.
& \textcolor{vred}{38.0} & \textcolor{vred}{48.6} & \textcolor{vred}{30.6} & \textcolor{vred}{33.1}
& \textcolor{vgreen}{24.0} & \textcolor{vred}{42.1} & \textcolor{vgreen}{49.5} & \textcolor{vgreen}{43.9}
& \textcolor{vred}{40.0} & \textcolor{vred}{37.8} & \textcolor{vred}{35.7} & \textcolor{vred}{38.8} \\

\bottomrule
\end{tabular}%
}

\vspace{0.8em}

\begin{tabular}{l|rrrr|rrrr}
\toprule
& \multicolumn{4}{c}{LLaVA-OV-7B}
& \multicolumn{4}{c}{Alpamayo} \\
\cmidrule(lr){2-5}
\cmidrule(lr){6-9}

Condition
& Acc  $\uparrow$ & AUROC  $\uparrow$ & ECE $\downarrow$& Brier $\downarrow$
& Acc  $\uparrow$ & AUROC $\uparrow$  & ECE$\downarrow$ & Brier $\downarrow$\\
\midrule

Clean
& 40.0 & 74.2 & 31.3 & 32.9
& 28.9 & 34.6 & 25.6 & 29.1 \\

Glare
& 40.0 & \textcolor{vgreen}{76.2} & \textcolor{vgreen}{31.3} & \textcolor{vgreen}{32.7}
& \textcolor{vgreen}{29.8} & \textcolor{vgreen}{39.6} & \textcolor{vred}{26.5} & \textcolor{vred}{29.3} \\

Fog
& 40.0 & \textcolor{vred}{62.5} & \textcolor{vred}{32.7} & \textcolor{vred}{34.3}
& \textcolor{vred}{27.3} & \textcolor{vred}{33.9} & \textcolor{vred}{36.3} & \textcolor{vred}{31.9} \\

Motion Blur
& \textcolor{vred}{36.0} & \textcolor{vred}{57.1} & \textcolor{vred}{37.1} & \textcolor{vred}{34.2}
& \textcolor{vgreen}{29.5} & \textcolor{vred}{32.3} & \textcolor{vred}{29.5} & \textcolor{vred}{30.9} \\

Lens Occl.
& \textcolor{vred}{38.0} & \textcolor{vred}{61.8} & \textcolor{vred}{33.7} & \textcolor{vred}{34.8}
& \textcolor{vgreen}{32.6} & \textcolor{vred}{30.8} & \textcolor{vred}{29.5} & \textcolor{vred}{31.0} \\

\bottomrule
\end{tabular}

\end{table}
Across NuScenes-QA-mini Figure~\ref{fig:vea_nuscenes}, STRIDE-QA Bench Figure~\ref{fig:vea_stride}, and OSR Figure~\ref{fig:vea_osr}, $\mathrm{V}{\scriptstyle \mathrm{EA}}$ does not produce a uniform change in reliability. Its effect depends on the model and task, and gains in accuracy or calibration are not always matched by better AUROC$_{\mathrm{VAUQ}}$. This supports the same overall finding that $\mathrm{V}{\scriptstyle \mathrm{EA}}$ should be assessed across both performance and confidence reliability rather than using a single metric.
\begin{figure}[h]
    \centering
    \input{vea_nuscenes}
    \caption{Effect of $\mathrm{V}{\scriptstyle \mathrm{EA}}$ on Accuracy, AUROC$_{\mathrm{VAUQ}}$, ECE, and Brier score on NuScenes-QA-mini under clean and corrupted conditions.}
    \label{fig:vea_nuscenes}
\end{figure}

\begin{figure}[h]
    \centering
    \input{vea_osr}
    \caption{Effect of $\mathrm{V}{\scriptstyle \mathrm{EA}}$ on Accuracy, AUROC$_{\mathrm{VAUQ}}$, ECE, and Brier score on OSR under clean and corrupted conditions.}
    \label{fig:vea_osr}
\end{figure}

\begin{figure}[h]
    \centering
    \input{vea_stride}
    \caption{Effect of $\mathrm{V}{\scriptstyle \mathrm{EA}}$ on Accuracy, AUROC$_{\mathrm{VAUQ}}$, ECE, and Brier score on STRIDE-QA under clean and corrupted conditions.}
    \label{fig:vea_stride}
\end{figure}
\section{$\mathrm{V}{\scriptstyle \mathrm{EA}}$ Computation Time}

We also measure the additional evidence extraction time introduced by $\mathrm{V}{\scriptstyle \mathrm{EA}}$, as shown in Figure~\ref{fig:vea-overhead-only}. For DrivingVQA, NuScenes-QA-mini, and OSR, the added time is generally below one second across models. The overhead is higher on STRIDE-QA Bench, reaching about 1.3 seconds for Qwen3.5-9B and 3 seconds for DriveFusionQA-4B. This shows that $\mathrm{V}{\scriptstyle \mathrm{EA}}$ can add only a small cost in many settings, but the cost is not consistent across models and tasks. For real time deployment, this additional latency is important because $\mathrm{V}{\scriptstyle \mathrm{EA}}$ is added on top of the original model inference time. Therefore, while the current results show that $\mathrm{V}{\scriptstyle \mathrm{EA}}$ can improve performance and reliability in several settings, reducing its computation cost would also be necessary for applications with strict real time requirements.
\input{latency}

\section{Implementation Details}
Table~\ref{tab:impl-details} lists the main implementation settings, and Table~\ref{tab:vea-layers} shows the $\mathrm{V}{\scriptstyle \mathrm{EA}}$ layers selected for each model.
\begin{table}[htbp]
\centering
\small
\caption{Hardware, generation, VAUQ hyperparameters, and evaluation sample sizes.}
\label{tab:impl-details}
\begin{tabular}{ll}
\toprule
\textbf{Parameter} & \textbf{Value} \\
\midrule
GPU & 2 $\times$ NVIDIA GeForce RTX 4090 (24\,GB each) \\
\midrule
\multicolumn{2}{l}{\textit{Generation}} \\
\texttt{max\_new\_tokens} & 128 (DrivingVQA, NuScenes-QA-mini, STRIDE-QA Bench); 640 (OSR); 512 ($\mathrm{V}{\scriptstyle \mathrm{EA}}$ DrivingVQA run) \\
\midrule
\multicolumn{2}{l}{\textit{VAUQ parameters}} \\
VAUQ alpha & 2.0 \\
Masked units & top-$k$ image \emph{token positions} by mean attention weight \\
Hooked layer range & layers $[\,n/3,\ 2n/3\,]$ of the model's total layers $n$, by default \\
\midrule
\multicolumn{2}{l}{\textit{Sample sizes}} \\
DrivingVQA & 1{,}000 \\
NuScenes-QA-mini & 1{,}117 \\
OSR & 50 (full dataset) \\
STRIDE-QA Bench & 1{,}636 total (\texttt{--limit 1000} used for most runs) \\
\bottomrule
\end{tabular}
\end{table}

\begin{table}[h]
\centering
\small
\caption{$\mathrm{V}{\scriptstyle \mathrm{EA}}$ steering layers selected per model (from profiling, top 10\% of layers by score, minimum 1). Alpamayo-1.5-10B uses a separate $\mathrm{V}{\scriptstyle \mathrm{EA}}$ implementation not covered by this profiling step.}
\label{tab:vea-layers}
\begin{tabular}{lccl}
\toprule
\textbf{Model} & \textbf{Total layers} & \textbf{\# Selected} & \textbf{Selected layer indices} \\
\midrule
Qwen3.5-9B       & 32 & 1 & 23 \\
DriveFusionQA-4B & 36 & 4 & 25, 27, 29, 31 \\
Gemma4-E4B       & 42 & 5 & 29, 32, 35, 38, 41 \\
LLaVA-OV-7B      & 28 & 3 & 25, 26, 27 \\
Alpamayo         & 36 & 4 & 0, 2, 4, 13 \\
\bottomrule
\end{tabular}
\end{table}

\end{document}

%% file: line_vea.tex
\definecolor{vblue}{RGB}{31,119,180}
\definecolor{vorange}{RGB}{255,127,14}
\definecolor{vgreen}{RGB}{0,140,70}
\definecolor{vred}{RGB}{190,30,45}
\definecolor{vgray}{RGB}{110,110,110}
\definecolor{vgrid}{RGB}{220,220,220}

\newcommand{\ChangeArrows}[2]{%
    \def\modearg{#1}%
    \def\upgoodarg{upgood}%
    \foreach \x/\yb/\yo in {#2} {%
        \pgfmathtruncatemacro{\dir}{abs(\yo-\yb) < 0.0001 ? 0 : (\yo>\yb ? 1 : -1)}%
        \pgfmathsetmacro{\xa}{\x+0.10}%
        \ifnum\dir=0
            \pgfmathsetmacro{\xl}{\x+0.04}%
            \pgfmathsetmacro{\xr}{\x+0.16}%
            \edef\tempcmd{\noexpand\draw[vgray,line width=1.0pt]
              (axis cs:\xl,\yb)--(axis cs:\xr,\yb);}%
            \tempcmd
        \else
            \ifx\modearg\upgoodarg
                \ifnum\dir=1 \def\arrowcolor{vgreen}\else \def\arrowcolor{vred}\fi
            \else
                \ifnum\dir=1 \def\arrowcolor{vred}\else \def\arrowcolor{vgreen}\fi
            \fi
            \edef\tempcmd{\noexpand\draw[-{Stealth[length=2.2mm,width=1.6mm]},draw=\arrowcolor,
              line width=0.9pt,shorten <=1pt,shorten >=1pt]
              (axis cs:\xa,\yb)--(axis cs:\xa,\yo);}%
            \tempcmd
        \fi
    }%
}

\begin{figure}[h]
\centering

\begin{tikzpicture}[baseline]
  \draw[vblue,solid,line width=1.3pt,mark=o,mark size=2.2pt] plot coordinates {(0,0) (0.8,0)};
  \node[anchor=west,font=\small] at (0.9,0) {Baseline};
  \draw[vorange,solid,line width=1.3pt,mark=square*,mark size=2.2pt] plot coordinates {(2.8,0) (3.6,0)};
  \node[anchor=west,font=\small] at (3.7,0) {VEA};
  \draw[-{Stealth[length=2.2mm,width=1.6mm]},vgreen,line width=1.0pt] (5.1,-0.15)--(5.1,0.15);
  \node[anchor=west,font=\small] at (5.25,0) {Favorable};
  \draw[-{Stealth[length=2.2mm,width=1.6mm]},vred,line width=1.0pt] (7.5,0.15)--(7.5,-0.15);
  \node[anchor=west,font=\small] at (7.65,0) {Unfavorable};
\end{tikzpicture}

\vspace{1mm}

\resizebox{0.99\linewidth}{!}{%
\begin{tikzpicture}
\begin{groupplot}[
    group style={group size=5 by 4,horizontal sep=0.75cm,vertical sep=0.60cm},
    width=4.4cm,height=3.8cm,scale only axis,
    xmin=0.5,xmax=5.5,xtick={1,2,3,4,5},
    xmajorgrids,ymajorgrids,
    major grid style={draw=vgrid,line width=0.5pt},
    axis line style={draw=black!70},tick style={draw=black!70},
    tick label style={font=\small},
    every axis plot/.append style={line width=1.3pt}
]

\nextgroupplot[title={Qwen3.5},title style={font=\normalsize\bfseries},ymin=52,ymax=61,ytick={52,54,56,58,60},xticklabels={},ylabel={Accuracy (\%)},ylabel style={font=\normalsize\bfseries}]
\addplot[vblue,solid,mark=o,mark size=2.2pt] coordinates {(1,55.80)(2,56.40)(3,55.30)(4,55.30)(5,53.80)};
\addplot[vorange,solid,mark=square*,mark size=2.2pt] coordinates {(1,59.00)(2,57.10)(3,57.50)(4,56.80)(5,58.80)};
\ChangeArrows{upgood}{1/55.80/59.00,2/56.40/57.10,3/55.30/57.50,4/55.30/56.80,5/53.80/58.80}

\nextgroupplot[title={DriveFusion},title style={font=\normalsize\bfseries},ymin=49.5,ymax=53,ytick={50,51,52,53},xticklabels={}]
\addplot[vblue,solid,mark=o,mark size=2.2pt] coordinates {(1,50.40)(2,50.90)(3,52.20)(4,50.70)(5,50.90)};
\addplot[vorange,solid,mark=square*,mark size=2.2pt] coordinates {(1,51.70)(2,51.00)(3,51.80)(4,50.90)(5,50.80)};
\ChangeArrows{upgood}{1/50.40/51.70,2/50.90/51.00,3/52.20/51.80,4/50.70/50.90,5/50.90/50.80}

\nextgroupplot[title={Gemma4},title style={font=\normalsize\bfseries},ymin=52.5,ymax=57,ytick={53,54,55,56,57},xticklabels={}]
\addplot[vblue,solid,mark=o,mark size=2.2pt] coordinates {(1,55.20)(2,54.70)(3,53.60)(4,53.30)(5,53.50)};
\addplot[vorange,solid,mark=square*,mark size=2.2pt] coordinates {(1,56.00)(2,54.10)(3,55.30)(4,56.30)(5,56.30)};
\ChangeArrows{upgood}{1/55.20/56.00,2/54.70/54.10,3/53.60/55.30,4/53.30/56.30,5/53.50/56.30}

\nextgroupplot[title={LLaVA-OV},title style={font=\normalsize\bfseries},ymin=51,ymax=55,ytick={51,52,53,54,55},xticklabels={}]
\addplot[vblue,solid,mark=o,mark size=2.2pt] coordinates {(1,53.80)(2,52.80)(3,51.60)(4,53.80)(5,53.20)};
\addplot[vorange,solid,mark=square*,mark size=2.2pt] coordinates {(1,54.30)(2,52.80)(3,53.00)(4,53.90)(5,52.80)};
\ChangeArrows{upgood}{1/53.80/54.30,2/52.80/52.80,3/51.60/53.00,4/53.80/53.90,5/53.20/52.80}

\nextgroupplot[title={Alpamayo},title style={font=\normalsize\bfseries},ymin=43,ymax=54,ytick={44,46,48,50,52,54},xticklabels={}]
\addplot[vblue,solid,mark=o,mark size=2.2pt] coordinates {(1,45.50)(2,45.00)(3,45.50)(4,44.40)(5,44.30)};
\addplot[vorange,solid,mark=square*,mark size=2.2pt] coordinates {(1,52.90)(2,51.70)(3,53.30)(4,51.90)(5,51.90)};
\ChangeArrows{upgood}{1/45.50/52.90,2/45.00/51.70,3/45.50/53.30,4/44.40/51.90,5/44.30/51.90}


\nextgroupplot[
    ymin=45, ymax=60,
    xticklabels={},
    ylabel={AUROC$_{\text{VAUQ}}$ (\%)},
    ylabel style={font=\normalsize\bfseries}
]

\addplot[black!40,dashed,line width=0.8pt,mark=none]
coordinates {(0.5,50)(5.5,50)};

\addplot[vblue,solid,mark=o,mark size=2.2pt] coordinates {
    (1,53.62)
    (2,54.56)
    (3,52.24)
    (4,53.50)
    (5,52.25)
};

\addplot[vorange,solid,mark=square*,mark size=2.2pt] coordinates {
    (1,52.10)
    (2,52.90)
    (3,53.48)
    (4,51.10)
    (5,52.18)
};

\ChangeArrows{upgood}{
    1/53.62/52.10,
    2/54.56/52.90,
    3/52.24/53.48,
    4/53.50/51.10,
    5/52.25/52.18
}

\nextgroupplot[
    ymin=45, ymax=60,
    xticklabels={},
    yticklabels={}
]

\addplot[black!40,dashed,line width=0.8pt,mark=none]
coordinates {(0.5,50)(5.5,50)};

\addplot[vblue,solid,mark=o,mark size=2.2pt] coordinates {
    (1,48.95)
    (2,48.66)
    (3,49.39)
    (4,51.36)
    (5,50.98)
};

\addplot[vorange,solid,mark=square*,mark size=2.2pt] coordinates {
    (1,52.51)
    (2,52.57)
    (3,53.16)
    (4,51.52)
    (5,52.60)
};

\ChangeArrows{upgood}{
    1/48.95/52.51,
    2/48.66/52.57,
    3/49.39/53.16,
    4/51.36/51.52,
    5/50.98/52.60
}

\nextgroupplot[
    ymin=45, ymax=60,
    xticklabels={},
    yticklabels={}
]

\addplot[black!40,dashed,line width=0.8pt,mark=none]
coordinates {(0.5,50)(5.5,50)};

\addplot[vblue,solid,mark=o,mark size=2.2pt] coordinates {
    (1,55.12)
    (2,52.33)
    (3,52.81)
    (4,49.57)
    (5,52.28)
};

\addplot[vorange,solid,mark=square*,mark size=2.2pt] coordinates {
    (1,51.23)
    (2,52.13)
    (3,52.46)
    (4,51.73)
    (5,50.90)
};

\ChangeArrows{upgood}{
    1/55.12/51.23,
    2/52.33/52.13,
    3/52.81/52.46,
    4/49.57/51.73,
    5/52.28/50.90
}

\nextgroupplot[
    ymin=45, ymax=60,
    xticklabels={},
    yticklabels={}
]

\addplot[black!40,dashed,line width=0.8pt,mark=none]
coordinates {(0.5,50)(5.5,50)};

\addplot[vblue,solid,mark=o,mark size=2.2pt] coordinates {
    (1,51.28)
    (2,52.41)
    (3,52.43)
    (4,50.84)
    (5,53.02)
};

\addplot[vorange,solid,mark=square*,mark size=2.2pt] coordinates {
    (1,54.14)
    (2,54.14)
    (3,52.42)
    (4,53.43)
    (5,54.01)
};

\ChangeArrows{upgood}{
    1/51.28/54.14,
    2/52.41/54.14,
    3/52.43/52.42,
    4/50.84/53.43,
    5/53.02/54.01
}

\nextgroupplot[
    ymin=45, ymax=60,
    xticklabels={},
    yticklabels={}
]

\addplot[black!40,dashed,line width=0.8pt,mark=none]
coordinates {(0.5,50)(5.5,50)};

\addplot[vblue,solid,mark=o,mark size=2.2pt] coordinates {
    (1,56.56)
    (2,56.13)
    (3,57.28)
    (4,56.15)
    (5,55.70)
};

\addplot[vorange,solid,mark=square*,mark size=2.2pt] coordinates {
    (1,52.34)
    (2,52.41)
    (3,48.72)
    (4,49.18)
    (5,49.98)
};

\ChangeArrows{upgood}{
    1/56.56/52.34,
    2/56.13/52.41,
    3/57.28/48.72,
    4/56.15/49.18,
    5/55.70/49.98
}

\nextgroupplot[ymin=16,ymax=23,ytick={16,18,20,22},xticklabels={},ylabel={ECE (\%)},ylabel style={font=\normalsize\bfseries}]
\addplot[vblue,solid,mark=o,mark size=2.2pt] coordinates {(1,21.51)(2,19.83)(3,21.87)(4,21.34)(5,22.65)};
\addplot[vorange,solid,mark=square*,mark size=2.2pt] coordinates {(1,17.44)(2,18.77)(3,18.75)(4,18.51)(5,16.95)};
\ChangeArrows{downgood}{1/21.51/17.44,2/19.83/18.77,3/21.87/18.75,4/21.34/18.51,5/22.65/16.95}

\nextgroupplot[ymin=9,ymax=14,ytick={10,11,12,13,14},xticklabels={}]
\addplot[vblue,solid,mark=o,mark size=2.2pt] coordinates {(1,13.50)(2,12.31)(3,11.15)(4,10.94)(5,11.18)};
\addplot[vorange,solid,mark=square*,mark size=2.2pt] coordinates {(1,10.52)(2,10.37)(3,9.85)(4,10.13)(5,10.75)};
\ChangeArrows{downgood}{1/13.50/10.52,2/12.31/10.37,3/11.15/9.85,4/10.94/10.13,5/11.18/10.75}

\nextgroupplot[ymin=15,ymax=22,ytick={16,18,20,22},xticklabels={}]
\addplot[vblue,solid,mark=o,mark size=2.2pt] coordinates {(1,19.50)(2,20.23)(3,21.34)(4,20.84)(5,21.88)};
\addplot[vorange,solid,mark=square*,mark size=2.2pt] coordinates {(1,17.05)(2,19.48)(3,17.84)(4,15.92)(5,16.28)};
\ChangeArrows{downgood}{1/19.50/17.05,2/20.23/19.48,3/21.34/17.84,4/20.84/15.92,5/21.88/16.28}

\nextgroupplot[ymin=19,ymax=24,ytick={20,21,22,23,24},xticklabels={}]
\addplot[vblue,solid,mark=o,mark size=2.2pt] coordinates {(1,21.93)(2,21.41)(3,22.38)(4,19.69)(5,21.10)};
\addplot[vorange,solid,mark=square*,mark size=2.2pt] coordinates {(1,22.80)(2,23.35)(3,22.86)(4,22.55)(5,23.42)};
\ChangeArrows{downgood}{1/21.93/22.80,2/21.41/23.35,3/22.38/22.86,4/19.69/22.55,5/21.10/23.42}

\nextgroupplot[ymin=23,ymax=31,ytick={24,26,28,30},xticklabels={}]
\addplot[vblue,solid,mark=o,mark size=2.2pt] coordinates {(1,23.98)(2,24.41)(3,24.37)(4,25.07)(5,25.16)};
\addplot[vorange,solid,mark=square*,mark size=2.2pt] coordinates {(1,30.50)(2,30.46)(3,29.88)(4,30.70)(5,30.81)};
\ChangeArrows{downgood}{1/23.98/30.50,2/24.41/30.46,3/24.37/29.88,4/25.07/30.70,5/25.16/30.81}

\nextgroupplot[ymin=27,ymax=30,ytick={27,28,29,30},xticklabels={Clean,Glare,Fog,Motion Blur,Lens Occlusion},x tick label style={rotate=30,anchor=east,font=\small},ylabel={Brier Score},ylabel style={font=\normalsize\bfseries}]
\addplot[vblue,solid,mark=o,mark size=2.2pt] coordinates {(1,29.28)(2,28.23)(3,29.52)(4,29.28)(5,30.07)};
\addplot[vorange,solid,mark=square*,mark size=2.2pt] coordinates {(1,27.24)(2,28.08)(3,27.98)(4,27.99)(5,27.06)};
\ChangeArrows{downgood}{1/29.28/27.24,2/28.23/28.08,3/29.52/27.98,4/29.28/27.99,5/30.07/27.06}

\nextgroupplot[ymin=26,ymax=27.5,ytick={26,26.5,27,27.5},xticklabels={Clean,Glare,Fog,Motion Blur,Lens Occlusion},x tick label style={rotate=30,anchor=east,font=\small}]
\addplot[vblue,solid,mark=o,mark size=2.2pt] coordinates {(1,27.15)(2,26.95)(3,26.59)(4,26.47)(5,26.72)};
\addplot[vorange,solid,mark=square*,mark size=2.2pt] coordinates {(1,26.14)(2,26.35)(3,26.32)(4,26.22)(5,26.36)};
\ChangeArrows{downgood}{1/27.15/26.14,2/26.95/26.35,3/26.59/26.32,4/26.47/26.22,5/26.72/26.36}

\nextgroupplot[ymin=27,ymax=30,ytick={27,28,29,30},xticklabels={Clean,Glare,Fog,Motion Blur,Lens Occlusion},x tick label style={rotate=30,anchor=east,font=\small}]
\addplot[vblue,solid,mark=o,mark size=2.2pt] coordinates {(1,28.37)(2,28.82)(3,29.54)(4,29.77)(5,29.74)};
\addplot[vorange,solid,mark=square*,mark size=2.2pt] coordinates {(1,27.58)(2,28.68)(3,28.06)(4,27.40)(5,27.63)};
\ChangeArrows{downgood}{1/28.37/27.58,2/28.82/28.68,3/29.54/28.06,4/29.77/27.40,5/29.74/27.63}

\nextgroupplot[ymin=29,ymax=31,ytick={29,29.5,30,30.5,31},xticklabels={Clean,Glare,Fog,Motion Blur,Lens Occlusion},x tick label style={rotate=30,anchor=east,font=\small}]
\addplot[vblue,solid,mark=o,mark size=2.2pt] coordinates {(1,30.21)(2,29.92)(3,30.63)(4,29.36)(5,29.76)};
\addplot[vorange,solid,mark=square*,mark size=2.2pt] coordinates {(1,30.60)(2,30.39)(3,30.83)(4,30.31)(5,30.62)};
\ChangeArrows{downgood}{1/30.21/30.60,2/29.92/30.39,3/30.63/30.83,4/29.36/30.31,5/29.76/30.62}

\nextgroupplot[ymin=30,ymax=37,ytick={30,32,34,36},xticklabels={Clean,Glare,Fog,Motion Blur,Lens Occlusion},x tick label style={rotate=30,anchor=east,font=\small}]
\addplot[vblue,solid,mark=o,mark size=2.2pt] coordinates {(1,30.80)(2,31.33)(3,31.11)(4,31.80)(5,31.56)};
\addplot[vorange,solid,mark=square*,mark size=2.2pt] coordinates {(1,35.37)(2,35.56)(3,35.50)(4,35.85)(5,36.05)};
\ChangeArrows{downgood}{1/30.80/35.37,2/31.33/35.56,3/31.11/35.50,4/31.80/35.85,5/31.56/36.05}

\end{groupplot}
\end{tikzpicture}%
}

\caption{Effect of VEA on Accuracy, AUROC$_{\mathrm{VAUQ}}$, ECE, and Brier score on DrivingVQA under clean and corrupted conditions.}
\label{fig:vea-lines}
\end{figure}
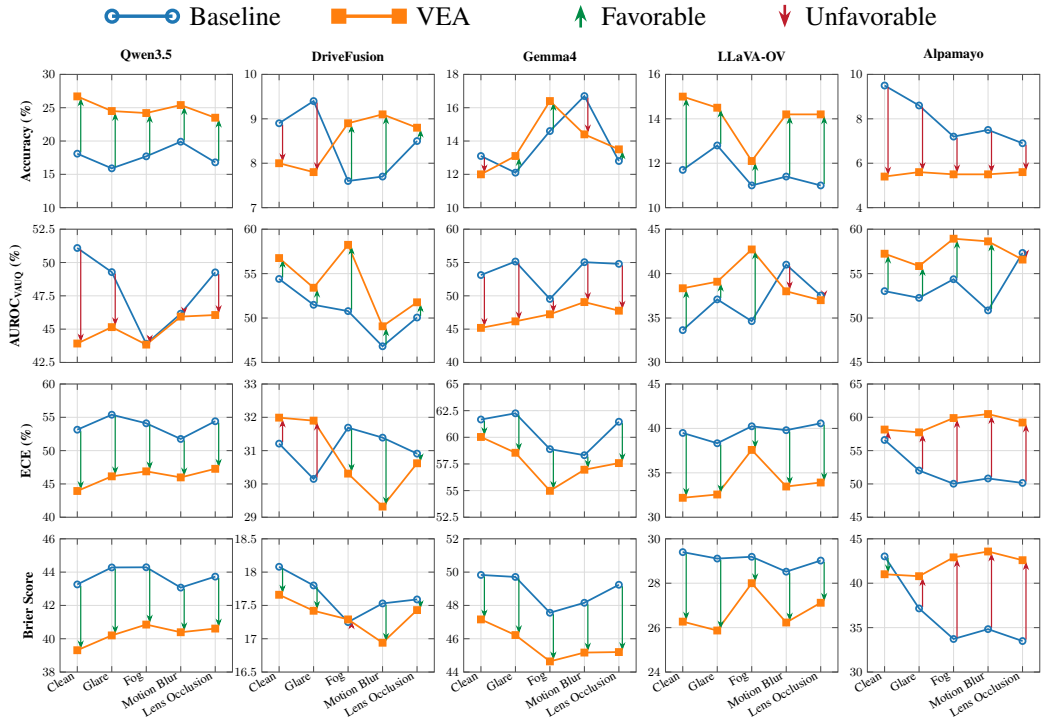

%% file: vea_nuscenes.tex
\begin{tikzpicture}[baseline]
  \draw[vblue,solid,line width=1.3pt,mark=o,mark size=2.2pt] plot coordinates {(0,0) (0.8,0)};
  \node[anchor=west,font=\normalsize] at (0.9,0) {Baseline};
  \draw[vorange,solid,line width=1.3pt,mark=square*,mark size=2.2pt] plot coordinates {(3.0,0) (3.8,0)};
  \node[anchor=west,font=\normalsize] at (3.9,0) {VEA};
  \draw[-{Stealth[length=2.2mm,width=1.6mm]},vgreen,line width=1.0pt] (6.2,-0.15) -- (6.2,0.15);
  \node[anchor=west,font=\normalsize] at (6.35,0) {Favorable};
  \draw[-{Stealth[length=2.2mm,width=1.6mm]},vred,line width=1.0pt] (8.9,0.15) -- (8.9,-0.15);
  \node[anchor=west,font=\normalsize] at (9.05,0) {Unfavorable};
\end{tikzpicture}

\vspace{1mm}

\resizebox{0.98\textwidth}{!}{%
\begin{tikzpicture}
\begin{groupplot}[
    group style={group size=5 by 4,horizontal sep=0.75cm,vertical sep=0.55cm},
    width=4.4cm,
    height=3.4cm,
    scale only axis,
    xmin=0.5,
    xmax=5.5,
    xtick={1,2,3,4,5},
    xmajorgrids,
    ymajorgrids,
    major grid style={draw=vgrid,line width=0.5pt},
    axis line style={draw=black!70},
    tick style={draw=black!70},
    tick label style={font=\small},
    every axis plot/.append style={line width=1.3pt},
]
\nextgroupplot[title={Qwen3.5}, title style={font=\normalsize\bfseries}, ymin=25.00, ymax=45.00, ytick={25.00,30.00,35.00,40.00,45.00}, xticklabels={}, ylabel={Accuracy (\%)}, ylabel style={font=\normalsize\bfseries}]

\addplot[vblue,solid,mark=o,mark size=2.2pt] coordinates {
    (1,42.08)
    (2,29.45)
    (3,40.56)
    (4,41.45)
    (5,40.91)
};

\addplot[vorange,solid,mark=square*,mark size=2.2pt] coordinates {
    (1,41.90)
    (2,34.83)
    (3,40.82)
    (4,41.18)
    (5,41.45)
};

\ChangeArrows{upgood}{
    1/42.08/41.90,
    2/29.45/34.83,
    3/40.56/40.82,
    4/41.45/41.18,
    5/40.91/41.45
}
\nextgroupplot[title={DriveFusion}, title style={font=\normalsize\bfseries}, ymin=34.00, ymax=38.00, ytick={34.00,35.00,36.00,37.00,38.00}, xticklabels={}]

\addplot[vblue,solid,mark=o,mark size=2.2pt] coordinates {
    (1,35.27)
    (2,36.44)
    (3,37.69)
    (4,36.26)
    (5,37.15)
};

\addplot[vorange,solid,mark=square*,mark size=2.2pt] coordinates {
    (1,36.53)
    (2,37.15)
    (3,36.26)
    (4,37.15)
    (5,37.69)
};

\ChangeArrows{upgood}{
    1/35.27/36.53,
    2/36.44/37.15,
    3/37.69/36.26,
    4/36.26/37.15,
    5/37.15/37.69
}
\nextgroupplot[title={Gemma4}, title style={font=\normalsize\bfseries}, ymin=25.00, ymax=40.00, ytick={25.00,30.00,35.00,40.00}, xticklabels={}]

\addplot[vblue,solid,mark=o,mark size=2.2pt] coordinates {
    (1,38.05)
    (2,29.81)
    (3,35.18)
    (4,34.65)
    (5,36.35)
};

\addplot[vorange,solid,mark=square*,mark size=2.2pt] coordinates {
    (1,37.24)
    (2,29.27)
    (3,35.09)
    (4,34.11)
    (5,36.44)
};

\ChangeArrows{upgood}{
    1/38.05/37.24,
    2/29.81/29.27,
    3/35.18/35.09,
    4/34.65/34.11,
    5/36.35/36.44
}
\nextgroupplot[title={LLaVA-OV}, title style={font=\normalsize\bfseries}, ymin=30.00, ymax=50.00, ytick={30.00,35.00,40.00,45.00,50.00}, xticklabels={}]

\addplot[vblue,solid,mark=o,mark size=2.2pt] coordinates {
    (1,44.49)
    (2,37.15)
    (3,42.35)
    (4,39.84)
    (5,44.67)
};

\addplot[vorange,solid,mark=square*,mark size=2.2pt] coordinates {
    (1,40.47)
    (2,34.74)
    (3,39.93)
    (4,38.68)
    (5,40.91)
};

\ChangeArrows{upgood}{
    1/44.49/40.47,
    2/37.15/34.74,
    3/42.35/39.93,
    4/39.84/38.68,
    5/44.67/40.91
}
\nextgroupplot[title={Alpamayo}, title style={font=\normalsize\bfseries}, ymin=40.00, ymax=60.00, ytick={40.00,45.00,50.00,55.00,60.00}, xticklabels={}]

\addplot[vblue,solid,mark=o,mark size=2.2pt] coordinates {
    (1,50.04)
    (2,45.12)
    (3,47.36)
    (4,44.23)
    (5,43.60)
};

\addplot[vorange,solid,mark=square*,mark size=2.2pt] coordinates {
    (1,58.01)
    (2,48.34)
    (3,51.75)
    (4,51.57)
    (5,52.10)
};

\ChangeArrows{upgood}{
    1/50.04/58.01,
    2/45.12/48.34,
    3/47.36/51.75,
    4/44.23/51.57,
    5/43.60/52.10
}
\nextgroupplot[ymin=45.00, ymax=70.00, ytick={45.00,50.00,55.00,60.00,65.00,70.00}, xticklabels={}, ylabel={AUROC$_{\text{VAUQ}}$ (\%)}, ylabel style={font=\normalsize\bfseries}]

\addplot[vblue,solid,mark=o,mark size=2.2pt] coordinates {
    (1,66.96)
    (2,60.04)
    (3,61.91)
    (4,61.79)
    (5,65.43)
};

\addplot[vorange,solid,mark=square*,mark size=2.2pt] coordinates {
    (1,65.89)
    (2,51.12)
    (3,60.86)
    (4,58.97)
    (5,64.66)
};

\ChangeArrows{upgood}{
    1/66.96/65.89,
    2/60.04/51.12,
    3/61.91/60.86,
    4/61.79/58.97,
    5/65.43/64.66
}
\nextgroupplot[ymin=50.00, ymax=65.00, ytick={50.00,55.00,60.00,65.00}, xticklabels={}]

\addplot[vblue,solid,mark=o,mark size=2.2pt] coordinates {
    (1,61.60)
    (2,54.94)
    (3,59.66)
    (4,58.28)
    (5,59.29)
};

\addplot[vorange,solid,mark=square*,mark size=2.2pt] coordinates {
    (1,63.17)
    (2,54.73)
    (3,60.20)
    (4,57.53)
    (5,60.37)
};

\ChangeArrows{upgood}{
    1/61.60/63.17,
    2/54.94/54.73,
    3/59.66/60.20,
    4/58.28/57.53,
    5/59.29/60.37
}
\nextgroupplot[ymin=50.00, ymax=65.00, ytick={50.00,55.00,60.00,65.00}, xticklabels={}]

\addplot[vblue,solid,mark=o,mark size=2.2pt] coordinates {
    (1,63.45)
    (2,55.06)
    (3,59.29)
    (4,57.77)
    (5,62.43)
};

\addplot[vorange,solid,mark=square*,mark size=2.2pt] coordinates {
    (1,63.86)
    (2,56.39)
    (3,57.45)
    (4,58.74)
    (5,63.89)
};

\ChangeArrows{upgood}{
    1/63.45/63.86,
    2/55.06/56.39,
    3/59.29/57.45,
    4/57.77/58.74,
    5/62.43/63.89
}
\nextgroupplot[ymin=50.00, ymax=70.00, ytick={50.00,55.00,60.00,65.00,70.00}, xticklabels={}]

\addplot[vblue,solid,mark=o,mark size=2.2pt] coordinates {
    (1,64.66)
    (2,52.11)
    (3,64.16)
    (4,60.13)
    (5,60.18)
};

\addplot[vorange,solid,mark=square*,mark size=2.2pt] coordinates {
    (1,62.02)
    (2,52.39)
    (3,61.94)
    (4,56.52)
    (5,59.51)
};

\ChangeArrows{upgood}{
    1/64.66/62.02,
    2/52.11/52.39,
    3/64.16/61.94,
    4/60.13/56.52,
    5/60.18/59.51
}
\nextgroupplot[ymin=45.00, ymax=65.00, ytick={45.00,50.00,55.00,60.00,65.00}, xticklabels={}]

\addplot[vblue,solid,mark=o,mark size=2.2pt] coordinates {
    (1,51.71)
    (2,55.16)
    (3,52.26)
    (4,54.29)
    (5,60.50)
};

\addplot[vorange,solid,mark=square*,mark size=2.2pt] coordinates {
    (1,50.78)
    (2,51.00)
    (3,50.17)
    (4,51.14)
    (5,52.56)
};

\ChangeArrows{upgood}{
    1/51.71/50.78,
    2/55.16/51.00,
    3/52.26/50.17,
    4/54.29/51.14,
    5/60.50/52.56
}
\nextgroupplot[ymin=45.00, ymax=70.00, ytick={45.00,50.00,55.00,60.00,65.00,70.00}, xticklabels={}, ylabel={ECE (\%)}, ylabel style={font=\normalsize\bfseries}]

\addplot[vblue,solid,mark=o,mark size=2.2pt] coordinates {
    (1,49.48)
    (2,63.41)
    (3,52.15)
    (4,51.85)
    (5,51.25)
};

\addplot[vorange,solid,mark=square*,mark size=2.2pt] coordinates {
    (1,49.75)
    (2,58.79)
    (3,51.90)
    (4,51.91)
    (5,50.56)
};

\ChangeArrows{downgood}{
    1/49.48/49.75,
    2/63.41/58.79,
    3/52.15/51.90,
    4/51.85/51.91,
    5/51.25/50.56
}
\nextgroupplot[ymin=20.00, ymax=25.00, ytick={20.00,21.00,22.00,23.00,24.00,25.00}, xticklabels={}]

\addplot[vblue,solid,mark=o,mark size=2.2pt] coordinates {
    (1,23.82)
    (2,21.95)
    (3,21.58)
    (4,23.11)
    (5,23.31)
};

\addplot[vorange,solid,mark=square*,mark size=2.2pt] coordinates {
    (1,21.03)
    (2,21.48)
    (3,22.64)
    (4,21.71)
    (5,22.21)
};

\ChangeArrows{downgood}{
    1/23.82/21.03,
    2/21.95/21.48,
    3/21.58/22.64,
    4/23.11/21.71,
    5/23.31/22.21
}
\nextgroupplot[ymin=50.00, ymax=65.00, ytick={50.00,55.00,60.00,65.00}, xticklabels={}]

\addplot[vblue,solid,mark=o,mark size=2.2pt] coordinates {
    (1,54.66)
    (2,63.31)
    (3,57.98)
    (4,59.51)
    (5,57.09)
};

\addplot[vorange,solid,mark=square*,mark size=2.2pt] coordinates {
    (1,55.08)
    (2,63.58)
    (3,57.93)
    (4,59.93)
    (5,56.55)
};

\ChangeArrows{downgood}{
    1/54.66/55.08,
    2/63.31/63.58,
    3/57.98/57.93,
    4/59.51/59.93,
    5/57.09/56.55
}
\nextgroupplot[ymin=25.00, ymax=45.00, ytick={25.00,30.00,35.00,40.00,45.00}, xticklabels={}]

\addplot[vblue,solid,mark=o,mark size=2.2pt] coordinates {
    (1,29.54)
    (2,40.43)
    (3,32.34)
    (4,33.36)
    (5,29.81)
};

\addplot[vorange,solid,mark=square*,mark size=2.2pt] coordinates {
    (1,30.37)
    (2,40.20)
    (3,31.35)
    (4,32.56)
    (5,30.16)
};

\ChangeArrows{downgood}{
    1/29.54/30.37,
    2/40.43/40.20,
    3/32.34/31.35,
    4/33.36/32.56,
    5/29.81/30.16
}
\nextgroupplot[ymin=10.00, ymax=25.00, ytick={10.00,15.00,20.00,25.00}, xticklabels={}]

\addplot[vblue,solid,mark=o,mark size=2.2pt] coordinates {
    (1,19.07)
    (2,21.93)
    (3,19.78)
    (4,20.64)
    (5,20.36)
};

\addplot[vorange,solid,mark=square*,mark size=2.2pt] coordinates {
    (1,13.38)
    (2,22.25)
    (3,18.77)
    (4,18.17)
    (5,18.12)
};

\ChangeArrows{downgood}{
    1/19.07/13.38,
    2/21.93/22.25,
    3/19.78/18.77,
    4/20.64/18.17,
    5/20.36/18.12
}
\nextgroupplot[ymin=45.00, ymax=65.00, ytick={45.00,50.00,55.00,60.00,65.00}, xticklabels={Clean,Glare,Fog,Motion Blur,Lens Occlusion}, x tick label style={rotate=30,anchor=east,font=\small}, ylabel={Brier Score}, ylabel style={font=\normalsize\bfseries}]

\addplot[vblue,solid,mark=o,mark size=2.2pt] coordinates {
    (1,47.46)
    (2,60.04)
    (3,50.50)
    (4,50.41)
    (5,49.43)
};

\addplot[vorange,solid,mark=square*,mark size=2.2pt] coordinates {
    (1,48.01)
    (2,57.50)
    (3,50.60)
    (4,50.69)
    (5,48.97)
};

\ChangeArrows{downgood}{
    1/47.46/48.01,
    2/60.04/57.50,
    3/50.50/50.60,
    4/50.41/50.69,
    5/49.43/48.97
}
\nextgroupplot[ymin=25.00, ymax=29.00, ytick={25.00,26.00,27.00,28.00,29.00}, xticklabels={Clean,Glare,Fog,Motion Blur,Lens Occlusion}, x tick label style={rotate=30,anchor=east,font=\small}]

\addplot[vblue,solid,mark=o,mark size=2.2pt] coordinates {
    (1,27.22)
    (2,28.55)
    (3,28.07)
    (4,28.32)
    (5,28.09)
};

\addplot[vorange,solid,mark=square*,mark size=2.2pt] coordinates {
    (1,26.00)
    (2,27.70)
    (3,27.67)
    (4,27.73)
    (5,27.71)
};

\ChangeArrows{downgood}{
    1/27.22/26.00,
    2/28.55/27.70,
    3/28.07/27.67,
    4/28.32/27.73,
    5/28.09/27.71
}
\nextgroupplot[ymin=50.00, ymax=65.00, ytick={50.00,55.00,60.00,65.00}, xticklabels={Clean,Glare,Fog,Motion Blur,Lens Occlusion}, x tick label style={rotate=30,anchor=east,font=\small}]

\addplot[vblue,solid,mark=o,mark size=2.2pt] coordinates {
    (1,51.66)
    (2,60.09)
    (3,54.99)
    (4,56.80)
    (5,53.96)
};

\addplot[vorange,solid,mark=square*,mark size=2.2pt] coordinates {
    (1,51.80)
    (2,60.20)
    (3,54.92)
    (4,57.21)
    (5,53.54)
};

\ChangeArrows{downgood}{
    1/51.66/51.80,
    2/60.09/60.20,
    3/54.99/54.92,
    4/56.80/57.21,
    5/53.96/53.54
}
\nextgroupplot[ymin=25.00, ymax=45.00, ytick={25.00,30.00,35.00,40.00,45.00}, xticklabels={Clean,Glare,Fog,Motion Blur,Lens Occlusion}, x tick label style={rotate=30,anchor=east,font=\small}]

\addplot[vblue,solid,mark=o,mark size=2.2pt] coordinates {
    (1,30.60)
    (2,40.58)
    (3,32.24)
    (4,33.59)
    (5,31.17)
};

\addplot[vorange,solid,mark=square*,mark size=2.2pt] coordinates {
    (1,32.14)
    (2,39.65)
    (3,32.67)
    (4,34.21)
    (5,32.11)
};

\ChangeArrows{downgood}{
    1/30.60/32.14,
    2/40.58/39.65,
    3/32.24/32.67,
    4/33.59/34.21,
    5/31.17/32.11
}
\nextgroupplot[ymin=24.00, ymax=32.00, ytick={24.00,26.00,28.00,30.00,32.00}, xticklabels={Clean,Glare,Fog,Motion Blur,Lens Occlusion}, x tick label style={rotate=30,anchor=east,font=\small}]

\addplot[vblue,solid,mark=o,mark size=2.2pt] coordinates {
    (1,28.58)
    (2,29.39)
    (3,28.76)
    (4,28.93)
    (5,28.07)
};

\addplot[vorange,solid,mark=square*,mark size=2.2pt] coordinates {
    (1,26.42)
    (2,30.17)
    (3,28.76)
    (4,28.50)
    (5,28.47)
};

\ChangeArrows{downgood}{
    1/28.58/26.42,
    2/29.39/30.17,
    3/28.76/28.76,
    4/28.93/28.50,
    5/28.07/28.47
}
\end{groupplot}
\end{tikzpicture}%
}

%% file: vea_osr.tex
\begin{tikzpicture}[baseline]
  \draw[vblue,solid,line width=1.3pt,mark=o,mark size=2.2pt] plot coordinates {(0,0) (0.8,0)};
  \node[anchor=west,font=\normalsize] at (0.9,0) {Baseline};
  \draw[vorange,solid,line width=1.3pt,mark=square*,mark size=2.2pt] plot coordinates {(3.0,0) (3.8,0)};
  \node[anchor=west,font=\normalsize] at (3.9,0) {VEA};
  \draw[-{Stealth[length=2.2mm,width=1.6mm]},vgreen,line width=1.0pt] (6.2,-0.15) -- (6.2,0.15);
  \node[anchor=west,font=\normalsize] at (6.35,0) {Favorable};
  \draw[-{Stealth[length=2.2mm,width=1.6mm]},vred,line width=1.0pt] (8.9,0.15) -- (8.9,-0.15);
  \node[anchor=west,font=\normalsize] at (9.05,0) {Unfavorable};
\end{tikzpicture}

\vspace{1mm}

\resizebox{0.98\textwidth}{!}{%
\begin{tikzpicture}
\begin{groupplot}[
    group style={group size=5 by 4,horizontal sep=0.75cm,vertical sep=0.55cm},
    width=4.4cm,
    height=3.4cm,
    scale only axis,
    xmin=0.5,
    xmax=5.5,
    xtick={1,2,3,4,5},
    xmajorgrids,
    ymajorgrids,
    major grid style={draw=vgrid,line width=0.5pt},
    axis line style={draw=black!70},
    tick style={draw=black!70},
    tick label style={font=\small},
    every axis plot/.append style={line width=1.3pt},
]
\nextgroupplot[title={Qwen3.5}, title style={font=\normalsize\bfseries}, ymin=30.00, ymax=55.00, ytick={30.00,35.00,40.00,45.00,50.00,55.00}, xticklabels={}, ylabel={Accuracy (\%)}, ylabel style={font=\normalsize\bfseries}]

\addplot[vblue,solid,mark=o,mark size=2.2pt] coordinates {
    (1,42.00)
    (2,50.00)
    (3,48.00)
    (4,36.00)
    (5,38.00)
};

\addplot[vorange,solid,mark=square*,mark size=2.2pt] coordinates {
    (1,50.00)
    (2,40.00)
    (3,46.00)
    (4,40.00)
    (5,46.00)
};

\ChangeArrows{upgood}{
    1/42.00/50.00,
    2/50.00/40.00,
    3/48.00/46.00,
    4/36.00/40.00,
    5/38.00/46.00
}
\nextgroupplot[title={DriveFusion}, title style={font=\normalsize\bfseries}, ymin=17.50, ymax=30.00, ytick={17.50,20.00,22.50,25.00,27.50,30.00}, xticklabels={}]

\addplot[vblue,solid,mark=o,mark size=2.2pt] coordinates {
    (1,20.00)
    (2,22.00)
    (3,28.00)
    (4,22.00)
    (5,24.00)
};

\addplot[vorange,solid,mark=square*,mark size=2.2pt] coordinates {
    (1,20.00)
    (2,24.00)
    (3,26.00)
    (4,22.00)
    (5,24.00)
};

\ChangeArrows{upgood}{
    1/20.00/20.00,
    2/22.00/24.00,
    3/28.00/26.00,
    4/22.00/22.00,
    5/24.00/24.00
}
\nextgroupplot[title={Gemma4}, title style={font=\normalsize\bfseries}, ymin=20.00, ymax=60.00, ytick={20.00,30.00,40.00,50.00,60.00}, xticklabels={}]

\addplot[vblue,solid,mark=o,mark size=2.2pt] coordinates {
    (1,42.00)
    (2,46.00)
    (3,42.00)
    (4,36.00)
    (5,40.00)
};

\addplot[vorange,solid,mark=square*,mark size=2.2pt] coordinates {
    (1,30.00)
    (2,32.00)
    (3,34.00)
    (4,38.00)
    (5,48.00)
};

\ChangeArrows{upgood}{
    1/42.00/30.00,
    2/46.00/32.00,
    3/42.00/34.00,
    4/36.00/38.00,
    5/40.00/48.00
}
\nextgroupplot[title={LLaVA-OV}, title style={font=\normalsize\bfseries}, ymin=32.00, ymax=42.00, ytick={32.00,34.00,36.00,38.00,40.00,42.00}, xticklabels={}]

\addplot[vblue,solid,mark=o,mark size=2.2pt] coordinates {
    (1,40.00)
    (2,40.00)
    (3,40.00)
    (4,36.00)
    (5,38.00)
};

\addplot[vorange,solid,mark=square*,mark size=2.2pt] coordinates {
    (1,36.00)
    (2,34.00)
    (3,34.00)
    (4,36.00)
    (5,40.00)
};

\ChangeArrows{upgood}{
    1/40.00/36.00,
    2/40.00/34.00,
    3/40.00/34.00,
    4/36.00/36.00,
    5/38.00/40.00
}
\nextgroupplot[title={Alpamayo}, title style={font=\normalsize\bfseries}, ymin=25.00, ymax=40.00, ytick={25.00,30.00,35.00,40.00}, xticklabels={}]

\addplot[vblue,solid,mark=o,mark size=2.2pt] coordinates {
    (1,28.89)
    (2,29.79)
    (3,27.27)
    (4,29.55)
    (5,32.61)
};

\addplot[vorange,solid,mark=square*,mark size=2.2pt] coordinates {
    (1,33.33)
    (2,29.79)
    (3,36.36)
    (4,34.09)
    (5,34.78)
};

\ChangeArrows{upgood}{
    1/28.89/33.33,
    2/29.79/29.79,
    3/27.27/36.36,
    4/29.55/34.09,
    5/32.61/34.78
}
\nextgroupplot[ymin=45.00, ymax=70.00, ytick={45.00,50.00,55.00,60.00,65.00,70.00}, xticklabels={}, ylabel={AUROC$_{\text{VAUQ}}$ (\%)}, ylabel style={font=\normalsize\bfseries}]

\addplot[vblue,solid,mark=o,mark size=2.2pt] coordinates {
    (1,53.37)
    (2,63.04)
    (3,57.37)
    (4,63.19)
    (5,48.56)
};

\addplot[vorange,solid,mark=square*,mark size=2.2pt] coordinates {
    (1,60.96)
    (2,63.50)
    (3,49.11)
    (4,59.00)
    (5,63.45)
};

\ChangeArrows{upgood}{
    1/53.37/60.96,
    2/63.04/63.50,
    3/57.37/49.11,
    4/63.19/59.00,
    5/48.56/63.45
}
\nextgroupplot[ymin=30.00, ymax=55.00, ytick={30.00,35.00,40.00,45.00,50.00,55.00}, xticklabels={}]

\addplot[vblue,solid,mark=o,mark size=2.2pt] coordinates {
    (1,44.50)
    (2,44.99)
    (3,42.86)
    (4,36.13)
    (5,42.11)
};

\addplot[vorange,solid,mark=square*,mark size=2.2pt] coordinates {
    (1,44.00)
    (2,39.91)
    (3,43.87)
    (4,51.98)
    (5,44.08)
};

\ChangeArrows{upgood}{
    1/44.50/44.00,
    2/44.99/39.91,
    3/42.86/43.87,
    4/36.13/51.98,
    5/42.11/44.08
}
\nextgroupplot[ymin=30.00, ymax=60.00, ytick={30.00,40.00,50.00,60.00}, xticklabels={}]

\addplot[vblue,solid,mark=o,mark size=2.2pt] coordinates {
    (1,56.65)
    (2,42.51)
    (3,41.22)
    (4,38.19)
    (5,37.83)
};

\addplot[vorange,solid,mark=square*,mark size=2.2pt] coordinates {
    (1,46.67)
    (2,37.50)
    (3,48.84)
    (4,36.33)
    (5,36.54)
};

\ChangeArrows{upgood}{
    1/56.65/46.67,
    2/42.51/37.50,
    3/41.22/48.84,
    4/38.19/36.33,
    5/37.83/36.54
}
\nextgroupplot[ymin=50.00, ymax=80.00, ytick={50.00,60.00,70.00,80.00}, xticklabels={}]

\addplot[vblue,solid,mark=o,mark size=2.2pt] coordinates {
    (1,74.17)
    (2,76.17)
    (3,62.50)
    (4,57.12)
    (5,61.80)
};

\addplot[vorange,solid,mark=square*,mark size=2.2pt] coordinates {
    (1,68.23)
    (2,60.96)
    (3,53.30)
    (4,60.59)
    (5,62.83)
};

\ChangeArrows{upgood}{
    1/74.17/68.23,
    2/76.17/60.96,
    3/62.50/53.30,
    4/57.12/60.59,
    5/61.80/62.83
}
\nextgroupplot[ymin=25.00, ymax=50.00, ytick={25.00,30.00,35.00,40.00,45.00,50.00}, xticklabels={}]

\addplot[vblue,solid,mark=o,mark size=2.2pt] coordinates {
    (1,34.62)
    (2,39.61)
    (3,33.85)
    (4,32.26)
    (5,30.75)
};

\addplot[vorange,solid,mark=square*,mark size=2.2pt] coordinates {
    (1,42.67)
    (2,45.89)
    (3,42.63)
    (4,40.69)
    (5,37.29)
};

\ChangeArrows{upgood}{
    1/34.62/42.67,
    2/39.61/45.89,
    3/33.85/42.63,
    4/32.26/40.69,
    5/30.75/37.29
}
\nextgroupplot[ymin=15.00, ymax=40.00, ytick={15.00,20.00,25.00,30.00,35.00,40.00}, xticklabels={}, ylabel={ECE (\%)}, ylabel style={font=\normalsize\bfseries}]

\addplot[vblue,solid,mark=o,mark size=2.2pt] coordinates {
    (1,27.36)
    (2,19.28)
    (3,21.81)
    (4,33.46)
    (5,30.64)
};

\addplot[vorange,solid,mark=square*,mark size=2.2pt] coordinates {
    (1,20.84)
    (2,28.72)
    (3,22.71)
    (4,28.82)
    (5,22.37)
};

\ChangeArrows{downgood}{
    1/27.36/20.84,
    2/19.28/28.72,
    3/21.81/22.71,
    4/33.46/28.82,
    5/30.64/22.37
}
\nextgroupplot[ymin=40.00, ymax=60.00, ytick={40.00,45.00,50.00,55.00,60.00}, xticklabels={}]

\addplot[vblue,solid,mark=o,mark size=2.2pt] coordinates {
    (1,54.56)
    (2,52.73)
    (3,45.80)
    (4,50.89)
    (5,49.49)
};

\addplot[vorange,solid,mark=square*,mark size=2.2pt] coordinates {
    (1,56.74)
    (2,52.67)
    (3,49.45)
    (4,53.93)
    (5,52.05)
};

\ChangeArrows{downgood}{
    1/54.56/56.74,
    2/52.73/52.67,
    3/45.80/49.45,
    4/50.89/53.93,
    5/49.49/52.05
}
\nextgroupplot[ymin=20.00, ymax=50.00, ytick={20.00,30.00,40.00,50.00}, xticklabels={}]

\addplot[vblue,solid,mark=o,mark size=2.2pt] coordinates {
    (1,34.44)
    (2,31.16)
    (3,34.75)
    (4,40.94)
    (5,35.68)
};

\addplot[vorange,solid,mark=square*,mark size=2.2pt] coordinates {
    (1,46.26)
    (2,44.34)
    (3,41.99)
    (4,38.10)
    (5,28.73)
};

\ChangeArrows{downgood}{
    1/34.44/46.26,
    2/31.16/44.34,
    3/34.75/41.99,
    4/40.94/38.10,
    5/35.68/28.73
}
\nextgroupplot[ymin=30.00, ymax=40.00, ytick={30.00,32.50,35.00,37.50,40.00}, xticklabels={}]

\addplot[vblue,solid,mark=o,mark size=2.2pt] coordinates {
    (1,31.31)
    (2,31.28)
    (3,32.68)
    (4,37.09)
    (5,33.66)
};

\addplot[vorange,solid,mark=square*,mark size=2.2pt] coordinates {
    (1,34.84)
    (2,36.50)
    (3,38.78)
    (4,35.93)
    (5,33.38)
};

\ChangeArrows{downgood}{
    1/31.31/34.84,
    2/31.28/36.50,
    3/32.68/38.78,
    4/37.09/35.93,
    5/33.66/33.38
}
\nextgroupplot[ymin=20.00, ymax=40.00, ytick={20.00,25.00,30.00,35.00,40.00}, xticklabels={}]

\addplot[vblue,solid,mark=o,mark size=2.2pt] coordinates {
    (1,25.62)
    (2,26.46)
    (3,36.28)
    (4,29.50)
    (5,29.46)
};

\addplot[vorange,solid,mark=square*,mark size=2.2pt] coordinates {
    (1,32.49)
    (2,27.83)
    (3,28.76)
    (4,24.62)
    (5,29.91)
};

\ChangeArrows{downgood}{
    1/25.62/32.49,
    2/26.46/27.83,
    3/36.28/28.76,
    4/29.50/24.62,
    5/29.46/29.91
}
\nextgroupplot[ymin=26.00, ymax=34.00, ytick={26.00,28.00,30.00,32.00,34.00}, xticklabels={Clean,Glare,Fog,Motion Blur,Lens Occlusion}, x tick label style={rotate=30,anchor=east,font=\small}, ylabel={Brier Score}, ylabel style={font=\normalsize\bfseries}]

\addplot[vblue,solid,mark=o,mark size=2.2pt] coordinates {
    (1,32.13)
    (2,27.80)
    (3,29.12)
    (4,33.09)
    (5,33.10)
};

\addplot[vorange,solid,mark=square*,mark size=2.2pt] coordinates {
    (1,28.70)
    (2,31.61)
    (3,30.69)
    (4,32.27)
    (5,29.10)
};

\ChangeArrows{downgood}{
    1/32.13/28.70,
    2/27.80/31.61,
    3/29.12/30.69,
    4/33.09/32.27,
    5/33.10/29.10
}
\nextgroupplot[ymin=42.00, ymax=50.00, ytick={42.00,44.00,46.00,48.00,50.00}, xticklabels={Clean,Glare,Fog,Motion Blur,Lens Occlusion}, x tick label style={rotate=30,anchor=east,font=\small}]

\addplot[vblue,solid,mark=o,mark size=2.2pt] coordinates {
    (1,46.92)
    (2,46.24)
    (3,43.11)
    (4,44.21)
    (5,43.87)
};

\addplot[vorange,solid,mark=square*,mark size=2.2pt] coordinates {
    (1,48.69)
    (2,46.87)
    (3,44.67)
    (4,46.66)
    (5,46.45)
};

\ChangeArrows{downgood}{
    1/46.92/48.69,
    2/46.24/46.87,
    3/43.11/44.67,
    4/44.21/46.66,
    5/43.87/46.45
}
\nextgroupplot[ymin=30.00, ymax=50.00, ytick={30.00,35.00,40.00,45.00,50.00}, xticklabels={Clean,Glare,Fog,Motion Blur,Lens Occlusion}, x tick label style={rotate=30,anchor=east,font=\small}]

\addplot[vblue,solid,mark=o,mark size=2.2pt] coordinates {
    (1,36.75)
    (2,35.94)
    (3,37.97)
    (4,40.94)
    (5,38.77)
};

\addplot[vorange,solid,mark=square*,mark size=2.2pt] coordinates {
    (1,43.90)
    (2,43.24)
    (3,41.29)
    (4,39.63)
    (5,34.62)
};

\ChangeArrows{downgood}{
    1/36.75/43.90,
    2/35.94/43.24,
    3/37.97/41.29,
    4/40.94/39.63,
    5/38.77/34.62
}
\nextgroupplot[ymin=32.00, ymax=38.00, ytick={32.00,34.00,36.00,38.00}, xticklabels={Clean,Glare,Fog,Motion Blur,Lens Occlusion}, x tick label style={rotate=30,anchor=east,font=\small}]

\addplot[vblue,solid,mark=o,mark size=2.2pt] coordinates {
    (1,32.88)
    (2,32.69)
    (3,34.29)
    (4,34.22)
    (5,34.80)
};

\addplot[vorange,solid,mark=square*,mark size=2.2pt] coordinates {
    (1,33.44)
    (2,34.93)
    (3,36.26)
    (4,34.56)
    (5,32.92)
};

\ChangeArrows{downgood}{
    1/32.88/33.44,
    2/32.69/34.93,
    3/34.29/36.26,
    4/34.22/34.56,
    5/34.80/32.92
}
\nextgroupplot[ymin=28.00, ymax=33.00, ytick={28.00,29.00,30.00,31.00,32.00,33.00}, xticklabels={Clean,Glare,Fog,Motion Blur,Lens Occlusion}, x tick label style={rotate=30,anchor=east,font=\small}]

\addplot[vblue,solid,mark=o,mark size=2.2pt] coordinates {
    (1,29.13)
    (2,29.25)
    (3,31.93)
    (4,30.91)
    (5,31.03)
};

\addplot[vorange,solid,mark=square*,mark size=2.2pt] coordinates {
    (1,30.99)
    (2,28.84)
    (3,30.29)
    (4,29.49)
    (5,31.09)
};

\ChangeArrows{downgood}{
    1/29.13/30.99,
    2/29.25/28.84,
    3/31.93/30.29,
    4/30.91/29.49,
    5/31.03/31.09
}
\end{groupplot}
\end{tikzpicture}%
}

%% file: vea_stride.tex
\begin{tikzpicture}[baseline]
  \draw[vblue,solid,line width=1.3pt,mark=o,mark size=2.2pt] plot coordinates {(0,0) (0.8,0)};
  \node[anchor=west,font=\normalsize] at (0.9,0) {Baseline};
  \draw[vorange,solid,line width=1.3pt,mark=square*,mark size=2.2pt] plot coordinates {(3.0,0) (3.8,0)};
  \node[anchor=west,font=\normalsize] at (3.9,0) {VEA};
  \draw[-{Stealth[length=2.2mm,width=1.6mm]},vgreen,line width=1.0pt] (6.2,-0.15) -- (6.2,0.15);
  \node[anchor=west,font=\normalsize] at (6.35,0) {Favorable};
  \draw[-{Stealth[length=2.2mm,width=1.6mm]},vred,line width=1.0pt] (8.9,0.15) -- (8.9,-0.15);
  \node[anchor=west,font=\normalsize] at (9.05,0) {Unfavorable};
\end{tikzpicture}

\vspace{1mm}

\resizebox{0.98\textwidth}{!}{%
\begin{tikzpicture}
\begin{groupplot}[
    group style={group size=5 by 4,horizontal sep=0.75cm,vertical sep=0.55cm},
    width=4.4cm,
    height=3.4cm,
    scale only axis,
    xmin=0.5,
    xmax=5.5,
    xtick={1,2,3,4,5},
    xmajorgrids,
    ymajorgrids,
    major grid style={draw=vgrid,line width=0.5pt},
    axis line style={draw=black!70},
    tick style={draw=black!70},
    tick label style={font=\small},
    every axis plot/.append style={line width=1.3pt},
]
\nextgroupplot[title={Qwen3.5}, title style={font=\normalsize\bfseries}, ymin=10.00, ymax=30.00, ytick={10.00,15.00,20.00,25.00,30.00}, xticklabels={}, ylabel={Accuracy (\%)}, ylabel style={font=\normalsize\bfseries}]

\addplot[vblue,solid,mark=o,mark size=2.2pt] coordinates {
    (1,18.10)
    (2,15.90)
    (3,17.70)
    (4,19.90)
    (5,16.80)
};

\addplot[vorange,solid,mark=square*,mark size=2.2pt] coordinates {
    (1,26.70)
    (2,24.50)
    (3,24.20)
    (4,25.40)
    (5,23.50)
};

\ChangeArrows{upgood}{
    1/18.10/26.70,
    2/15.90/24.50,
    3/17.70/24.20,
    4/19.90/25.40,
    5/16.80/23.50
}
\nextgroupplot[title={DriveFusion}, title style={font=\normalsize\bfseries}, ymin=7.00, ymax=10.00, ytick={7.00,8.00,9.00,10.00}, xticklabels={}]

\addplot[vblue,solid,mark=o,mark size=2.2pt] coordinates {
    (1,8.90)
    (2,9.40)
    (3,7.60)
    (4,7.70)
    (5,8.50)
};

\addplot[vorange,solid,mark=square*,mark size=2.2pt] coordinates {
    (1,8.00)
    (2,7.80)
    (3,8.90)
    (4,9.10)
    (5,8.80)
};

\ChangeArrows{upgood}{
    1/8.90/8.00,
    2/9.40/7.80,
    3/7.60/8.90,
    4/7.70/9.10,
    5/8.50/8.80
}
\nextgroupplot[title={Gemma4}, title style={font=\normalsize\bfseries}, ymin=10.00, ymax=18.00, ytick={10.00,12.00,14.00,16.00,18.00}, xticklabels={}]

\addplot[vblue,solid,mark=o,mark size=2.2pt] coordinates {
    (1,13.10)
    (2,12.10)
    (3,14.60)
    (4,16.70)
    (5,12.80)
};

\addplot[vorange,solid,mark=square*,mark size=2.2pt] coordinates {
    (1,12.00)
    (2,13.10)
    (3,16.40)
    (4,14.40)
    (5,13.50)
};

\ChangeArrows{upgood}{
    1/13.10/12.00,
    2/12.10/13.10,
    3/14.60/16.40,
    4/16.70/14.40,
    5/12.80/13.50
}
\nextgroupplot[title={LLaVA-OV}, title style={font=\normalsize\bfseries}, ymin=10.00, ymax=16.00, ytick={10.00,12.00,14.00,16.00}, xticklabels={}]

\addplot[vblue,solid,mark=o,mark size=2.2pt] coordinates {
    (1,11.70)
    (2,12.80)
    (3,11.00)
    (4,11.40)
    (5,11.00)
};

\addplot[vorange,solid,mark=square*,mark size=2.2pt] coordinates {
    (1,15.00)
    (2,14.50)
    (3,12.10)
    (4,14.20)
    (5,14.20)
};

\ChangeArrows{upgood}{
    1/11.70/15.00,
    2/12.80/14.50,
    3/11.00/12.10,
    4/11.40/14.20,
    5/11.00/14.20
}
\nextgroupplot[title={Alpamayo}, title style={font=\normalsize\bfseries}, ymin=4.00, ymax=10.00, ytick={4.00,6.00,8.00,10.00}, xticklabels={}]

\addplot[vblue,solid,mark=o,mark size=2.2pt] coordinates {
    (1,9.50)
    (2,8.60)
    (3,7.20)
    (4,7.50)
    (5,6.90)
};

\addplot[vorange,solid,mark=square*,mark size=2.2pt] coordinates {
    (1,5.40)
    (2,5.60)
    (3,5.50)
    (4,5.50)
    (5,5.60)
};

\ChangeArrows{upgood}{
    1/9.50/5.40,
    2/8.60/5.60,
    3/7.20/5.50,
    4/7.50/5.50,
    5/6.90/5.60
}
\nextgroupplot[ymin=42.50, ymax=52.50, ytick={42.50,45.00,47.50,50.00,52.50}, xticklabels={}, ylabel={AUROC$_{\text{VAUQ}}$ (\%)}, ylabel style={font=\normalsize\bfseries}]

\addplot[vblue,solid,mark=o,mark size=2.2pt] coordinates {
    (1,51.09)
    (2,49.28)
    (3,43.90)
    (4,46.16)
    (5,49.26)
};

\addplot[vorange,solid,mark=square*,mark size=2.2pt] coordinates {
    (1,43.92)
    (2,45.15)
    (3,43.84)
    (4,45.94)
    (5,46.06)
};

\ChangeArrows{upgood}{
    1/51.09/43.92,
    2/49.28/45.15,
    3/43.90/43.84,
    4/46.16/45.94,
    5/49.26/46.06
}
\nextgroupplot[ymin=45.00, ymax=60.00, ytick={45.00,50.00,55.00,60.00}, xticklabels={}]

\addplot[vblue,solid,mark=o,mark size=2.2pt] coordinates {
    (1,54.40)
    (2,51.49)
    (3,50.78)
    (4,46.82)
    (5,50.06)
};

\addplot[vorange,solid,mark=square*,mark size=2.2pt] coordinates {
    (1,56.75)
    (2,53.41)
    (3,58.24)
    (4,49.07)
    (5,51.77)
};

\ChangeArrows{upgood}{
    1/54.40/56.75,
    2/51.49/53.41,
    3/50.78/58.24,
    4/46.82/49.07,
    5/50.06/51.77
}
\nextgroupplot[ymin=40.00, ymax=60.00, ytick={40.00,45.00,50.00,55.00,60.00}, xticklabels={}]

\addplot[vblue,solid,mark=o,mark size=2.2pt] coordinates {
    (1,53.12)
    (2,55.15)
    (3,49.53)
    (4,55.05)
    (5,54.80)
};

\addplot[vorange,solid,mark=square*,mark size=2.2pt] coordinates {
    (1,45.19)
    (2,46.17)
    (3,47.23)
    (4,49.06)
    (5,47.77)
};

\ChangeArrows{upgood}{
    1/53.12/45.19,
    2/55.15/46.17,
    3/49.53/47.23,
    4/55.05/49.06,
    5/54.80/47.77
}
\nextgroupplot[ymin=30.00, ymax=45.00, ytick={30.00,35.00,40.00,45.00}, xticklabels={}]

\addplot[vblue,solid,mark=o,mark size=2.2pt] coordinates {
    (1,33.64)
    (2,37.10)
    (3,34.65)
    (4,41.01)
    (5,37.53)
};

\addplot[vorange,solid,mark=square*,mark size=2.2pt] coordinates {
    (1,38.35)
    (2,39.09)
    (3,42.72)
    (4,38.01)
    (5,37.01)
};

\ChangeArrows{upgood}{
    1/33.64/38.35,
    2/37.10/39.09,
    3/34.65/42.72,
    4/41.01/38.01,
    5/37.53/37.01
}
\nextgroupplot[ymin=45.00, ymax=60.00, ytick={45.00,50.00,55.00,60.00}, xticklabels={}]

\addplot[vblue,solid,mark=o,mark size=2.2pt] coordinates {
    (1,53.03)
    (2,52.27)
    (3,54.37)
    (4,50.86)
    (5,57.33)
};

\addplot[vorange,solid,mark=square*,mark size=2.2pt] coordinates {
    (1,57.24)
    (2,55.85)
    (3,58.93)
    (4,58.63)
    (5,56.58)
};

\ChangeArrows{upgood}{
    1/53.03/57.24,
    2/52.27/55.85,
    3/54.37/58.93,
    4/50.86/58.63,
    5/57.33/56.58
}
\nextgroupplot[ymin=40.00, ymax=60.00, ytick={40.00,45.00,50.00,55.00,60.00}, xticklabels={}, ylabel={ECE (\%)}, ylabel style={font=\normalsize\bfseries}]

\addplot[vblue,solid,mark=o,mark size=2.2pt] coordinates {
    (1,53.14)
    (2,55.39)
    (3,54.10)
    (4,51.75)
    (5,54.41)
};

\addplot[vorange,solid,mark=square*,mark size=2.2pt] coordinates {
    (1,43.94)
    (2,46.14)
    (3,46.89)
    (4,45.98)
    (5,47.27)
};

\ChangeArrows{downgood}{
    1/53.14/43.94,
    2/55.39/46.14,
    3/54.10/46.89,
    4/51.75/45.98,
    5/54.41/47.27
}
\nextgroupplot[ymin=29.00, ymax=33.00, ytick={29.00,30.00,31.00,32.00,33.00}, xticklabels={}]

\addplot[vblue,solid,mark=o,mark size=2.2pt] coordinates {
    (1,31.21)
    (2,30.15)
    (3,31.69)
    (4,31.39)
    (5,30.91)
};

\addplot[vorange,solid,mark=square*,mark size=2.2pt] coordinates {
    (1,31.99)
    (2,31.90)
    (3,30.31)
    (4,29.32)
    (5,30.62)
};

\ChangeArrows{downgood}{
    1/31.21/31.99,
    2/30.15/31.90,
    3/31.69/30.31,
    4/31.39/29.32,
    5/30.91/30.62
}
\nextgroupplot[ymin=52.50, ymax=65.00, ytick={52.50,55.00,57.50,60.00,62.50,65.00}, xticklabels={}]

\addplot[vblue,solid,mark=o,mark size=2.2pt] coordinates {
    (1,61.67)
    (2,62.25)
    (3,58.89)
    (4,58.32)
    (5,61.46)
};

\addplot[vorange,solid,mark=square*,mark size=2.2pt] coordinates {
    (1,60.03)
    (2,58.55)
    (3,54.98)
    (4,56.96)
    (5,57.58)
};

\ChangeArrows{downgood}{
    1/61.67/60.03,
    2/62.25/58.55,
    3/58.89/54.98,
    4/58.32/56.96,
    5/61.46/57.58
}
\nextgroupplot[ymin=30.00, ymax=45.00, ytick={30.00,35.00,40.00,45.00}, xticklabels={}]

\addplot[vblue,solid,mark=o,mark size=2.2pt] coordinates {
    (1,39.49)
    (2,38.33)
    (3,40.24)
    (4,39.80)
    (5,40.57)
};

\addplot[vorange,solid,mark=square*,mark size=2.2pt] coordinates {
    (1,32.19)
    (2,32.56)
    (3,37.58)
    (4,33.46)
    (5,33.91)
};

\ChangeArrows{downgood}{
    1/39.49/32.19,
    2/38.33/32.56,
    3/40.24/37.58,
    4/39.80/33.46,
    5/40.57/33.91
}
\nextgroupplot[ymin=45.00, ymax=65.00, ytick={45.00,50.00,55.00,60.00,65.00}, xticklabels={}]

\addplot[vblue,solid,mark=o,mark size=2.2pt] coordinates {
    (1,56.60)
    (2,52.01)
    (3,50.03)
    (4,50.82)
    (5,50.15)
};

\addplot[vorange,solid,mark=square*,mark size=2.2pt] coordinates {
    (1,58.16)
    (2,57.74)
    (3,59.89)
    (4,60.49)
    (5,59.22)
};

\ChangeArrows{downgood}{
    1/56.60/58.16,
    2/52.01/57.74,
    3/50.03/59.89,
    4/50.82/60.49,
    5/50.15/59.22
}
\nextgroupplot[ymin=38.00, ymax=46.00, ytick={38.00,40.00,42.00,44.00,46.00}, xticklabels={Clean,Glare,Fog,Motion Blur,Lens Occlusion}, x tick label style={rotate=30,anchor=east,font=\small}, ylabel={Brier Score}, ylabel style={font=\normalsize\bfseries}]

\addplot[vblue,solid,mark=o,mark size=2.2pt] coordinates {
    (1,43.26)
    (2,44.28)
    (3,44.29)
    (4,43.07)
    (5,43.73)
};

\addplot[vorange,solid,mark=square*,mark size=2.2pt] coordinates {
    (1,39.31)
    (2,40.20)
    (3,40.85)
    (4,40.39)
    (5,40.61)
};

\ChangeArrows{downgood}{
    1/43.26/39.31,
    2/44.28/40.20,
    3/44.29/40.85,
    4/43.07/40.39,
    5/43.73/40.61
}
\nextgroupplot[ymin=16.50, ymax=18.50, ytick={16.50,17.00,17.50,18.00,18.50}, xticklabels={Clean,Glare,Fog,Motion Blur,Lens Occlusion}, x tick label style={rotate=30,anchor=east,font=\small}]

\addplot[vblue,solid,mark=o,mark size=2.2pt] coordinates {
    (1,18.08)
    (2,17.80)
    (3,17.25)
    (4,17.53)
    (5,17.59)
};

\addplot[vorange,solid,mark=square*,mark size=2.2pt] coordinates {
    (1,17.66)
    (2,17.42)
    (3,17.29)
    (4,16.94)
    (5,17.43)
};

\ChangeArrows{downgood}{
    1/18.08/17.66,
    2/17.80/17.42,
    3/17.25/17.29,
    4/17.53/16.94,
    5/17.59/17.43
}
\nextgroupplot[ymin=44.00, ymax=52.00, ytick={44.00,46.00,48.00,50.00,52.00}, xticklabels={Clean,Glare,Fog,Motion Blur,Lens Occlusion}, x tick label style={rotate=30,anchor=east,font=\small}]

\addplot[vblue,solid,mark=o,mark size=2.2pt] coordinates {
    (1,49.83)
    (2,49.71)
    (3,47.56)
    (4,48.16)
    (5,49.24)
};

\addplot[vorange,solid,mark=square*,mark size=2.2pt] coordinates {
    (1,47.16)
    (2,46.22)
    (3,44.64)
    (4,45.17)
    (5,45.20)
};

\ChangeArrows{downgood}{
    1/49.83/47.16,
    2/49.71/46.22,
    3/47.56/44.64,
    4/48.16/45.17,
    5/49.24/45.20
}
\nextgroupplot[ymin=24.00, ymax=30.00, ytick={24.00,26.00,28.00,30.00}, xticklabels={Clean,Glare,Fog,Motion Blur,Lens Occlusion}, x tick label style={rotate=30,anchor=east,font=\small}]

\addplot[vblue,solid,mark=o,mark size=2.2pt] coordinates {
    (1,29.40)
    (2,29.11)
    (3,29.19)
    (4,28.52)
    (5,29.02)
};

\addplot[vorange,solid,mark=square*,mark size=2.2pt] coordinates {
    (1,26.27)
    (2,25.87)
    (3,28.00)
    (4,26.23)
    (5,27.12)
};

\ChangeArrows{downgood}{
    1/29.40/26.27,
    2/29.11/25.87,
    3/29.19/28.00,
    4/28.52/26.23,
    5/29.02/27.12
}
\nextgroupplot[ymin=30.00, ymax=45.00, ytick={30.00,35.00,40.00,45.00}, xticklabels={Clean,Glare,Fog,Motion Blur,Lens Occlusion}, x tick label style={rotate=30,anchor=east,font=\small}]

\addplot[vblue,solid,mark=o,mark size=2.2pt] coordinates {
    (1,43.02)
    (2,37.16)
    (3,33.72)
    (4,34.83)
    (5,33.49)
};

\addplot[vorange,solid,mark=square*,mark size=2.2pt] coordinates {
    (1,41.01)
    (2,40.79)
    (3,42.91)
    (4,43.57)
    (5,42.59)
};

\ChangeArrows{downgood}{
    1/43.02/41.01,
    2/37.16/40.79,
    3/33.72/42.91,
    4/34.83/43.57,
    5/33.49/42.59
}
\end{groupplot}
\end{tikzpicture}%
}

%% file: latency.tex
\definecolor{mQwen}{HTML}{2a78d6}
\definecolor{mDriveFusion}{HTML}{eb6834}
\definecolor{mGemma4}{HTML}{1baf7a}
\definecolor{mLLaVAOV}{HTML}{eda100}
\definecolor{mAlpamayo}{HTML}{e87ba4}

\begin{figure}[htbp]
\centering

\begin{tikzpicture}

\begin{groupplot}[
    group style={
        group size=1 by 1
    },
    ybar,
    /pgf/bar width=5pt,
    width=8.5cm,
    height=6cm,
    symbolic x coords={DVQA,NQA,OSR,STRIDE},
    xtick=data,
    x tick label style={font=\small},
    ymin=0,
    enlarge y limits={value=0.12,upper},
    enlarge x limits=0.18,
    tick label style={font=\normalsize},
    axis lines*=left,
    legend columns=5,
    legend style={
        font=\normalsize,
        draw=none,
        at={(0.5,-0.18)},
        anchor=north,
        /tikz/every even column/.append style={
            column sep=0.2cm
        }
    }
]

\nextgroupplot[
    title={},
    ylabel={VEA evidence-extraction time (s)}
]

\addplot[
    fill=mQwen,
    draw=white,
    line width=0.2pt
] coordinates {
    (DVQA,0.268)
    (NQA,0.116)
    (OSR,0.258)
    (STRIDE,1.279)
};

\addplot[
    fill=mDriveFusion,
    draw=white,
    line width=0.2pt
] coordinates {
    (DVQA,0.260)
    (NQA,0.162)
    (OSR,0.546)
    (STRIDE,3.063)
};

\addplot[
    fill=mGemma4,
    draw=white,
    line width=0.2pt
] coordinates {
    (DVQA,0.135)
    (NQA,0.798)
    (OSR,0.186)
    (STRIDE,0.399)
};

\addplot[
    fill=mLLaVAOV,
    draw=white,
    line width=0.2pt
] coordinates {
    (DVQA,0.061)
    (NQA,0.495)
    (OSR,0.141)
    (STRIDE,0.278)
};

\addplot[
    fill=mAlpamayo,
    draw=white,
    line width=0.2pt
] coordinates {
    (DVQA,0.076)
    (NQA,0.456)
    (OSR,0.162)
    (STRIDE,0.572)
};

\legend{
    Qwen3.5,
    DriveFusion,
    Gemma4,
    LLaVA-OV,
    Alpamayo
}

\end{groupplot}

\end{tikzpicture}

\caption{
Additional evidence extraction time introduced by VEA across models and datasets. Values show the average added time per sample.
}

\label{fig:vea-overhead-only}
\end{figure}
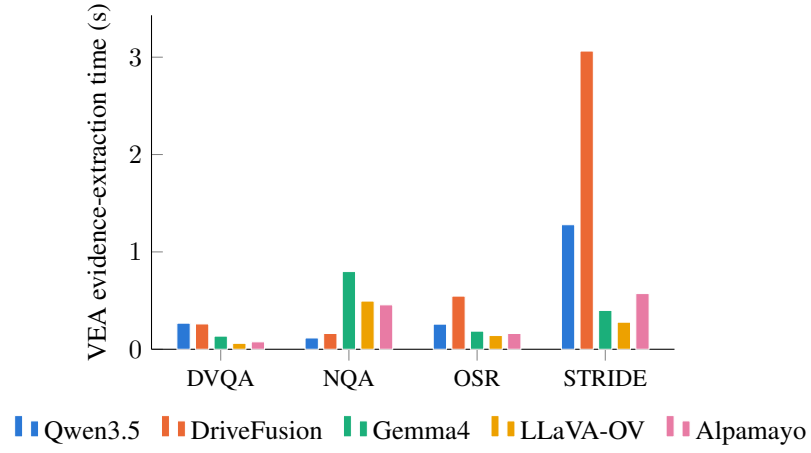